\documentclass[10pt,letterpaper]{article}
\usepackage[T1]{fontenc}
\usepackage[utf8]{inputenc}
\usepackage[margin=1in]{geometry}
\usepackage{times}
\usepackage{amsmath,amssymb}
\usepackage{graphicx}
\usepackage{booktabs,longtable,array,calc}
\usepackage{microtype}
\usepackage[authoryear,round]{natbib}
\usepackage{url}
\usepackage[hidelinks]{hyperref}

\makeatletter
\def\maxwidth{\ifdim\Gin@nat@width>\linewidth\linewidth\else\Gin@nat@width\fi}
\def\maxheight{\ifdim\Gin@nat@height>\textheight\textheight\else\Gin@nat@height\fi}
\makeatother
\setkeys{Gin}{width=\maxwidth,height=\maxheight,keepaspectratio}

\hypersetup{pdftitle={PINNForge: Execution-Grounded Evolutionary Design of Physics-Informed Neural Networks},pdfauthor={},pdfsubject={},pdfkeywords={}}

\title{PINNForge: Execution-Grounded Evolutionary Design of Physics-Informed Neural Networks}
\author{
Mingyang Yu$^{1}$,
Xu Yang$^{2}$,
Jun Zhang$^{3}$,
Jing Xu$^{1}$,
Keqian Li$^{2}$\\[0.5em]
$^{1}$Nankai University\\
$^{2}$East China Normal University\\
$^{3}$Hanyang University
}
\date{}

\begin{document}
\maketitle

\begin{abstract}
Physics-informed neural networks (PINNs) require coordinated choices across representation, architecture, sampling, constraints, loss construction, and optimization, yet effective configurations can vary substantially across partial differential equations (PDEs). Existing automated design methods search over these choices, but training outcomes are still used primarily to rank candidates rather than to inform subsequent designs. We develop PINNForge, an execution-grounded large language model (LLM)-driven evolutionary framework that treats observed PINN training behavior as a cross-generation design signal. PINNForge initializes diverse configurations from PDE-related prior knowledge, executes them, retains globally strong configurations, and feeds their execution evidence and accumulated run-level experience back to the LLM. The LLM then revises and recombines coupled design components or explores new combinations within a generate--execute--evaluate--evolve loop. Across 25 PDE benchmarks, PINNForge achieves the lowest mean MSE on 24 tasks among RoPINN, PINNsFormer, PINNsAgent, and PINNForge. Removing knowledge guidance, execution feedback, or evolutionary search increases the median task-wise MSE ratio to 3.74$\times$, 12.10$\times$, and 10.10$\times$, respectively, relative to full PINNForge.
\end{abstract}

\section{Introduction}\label{sec:introduction}

Physics-informed neural networks (PINNs) solve partial differential equations (PDEs) by embedding governing equations and physical constraints into neural-network training \citep{raissi2019pinn,toscano2025pikan,bai2025future,michaloglou2026review}. In practice, however, their accuracy depends on a set of strongly coupled design choices, including input representation, network architecture, collocation sampling, constraint handling, loss construction, and optimization \citep{krishnapriyan2021failure,rathore2024challenges,wang2021gradient,wang2022ntk,wu2023sampling,wang2024causal}. The dominant bottleneck can also differ substantially across PDEs: multiscale problems may demand more expressive representations, complex geometries may rely more heavily on sampling and constraint enforcement, and long-time systems may be limited by causality or optimization stability. Consequently, a configuration that performs well on one PDE may not transfer effectively to another.

This motivates automated PINN design. Existing work has explored neural architecture search, hyperparameter optimization, evolutionary computation, and LLM-based agents to reduce manual trial and error \citep{wang2023autopinn,wangzhong2024naspinn,kaplarevic2023optimal,zhangyang2024evo,carrillo2024mopinns,wuwu2025pinnsagent,he2026langpinn}. These approaches demonstrate that important PINN design choices can be automated, but training outcomes are still used primarily to rank candidates or diagnose isolated failures. In contrast, execution provides richer evidence---including PDE residuals, physical-constraint errors, loss and gradient behavior, stagnation, and numerical stability---that can reveal not only which candidate performs better, but also why a configuration underperforms and which aspects of the design may require revision.

The challenge is further compounded by interactions among PINN components. A representation bottleneck, for example, may require coordinated changes to architecture, sampling, loss balancing, and optimization rather than an isolated numerical adjustment. Automated design therefore calls for a search mechanism that reasons over complete configurations, preserves effective mechanisms, revises coupled components in response to observed failures, and explores beyond the current elite set.

Recent LLM-driven algorithm-discovery methods provide a natural foundation for this direction. FunSearch, Evolution of Heuristics (EoH), ReEvo, and LLaMEA embed LLMs within iterative generate--execute--evaluate--evolve loops, using strong candidates and execution outcomes to guide subsequent proposals \citep{romeraparedes2024funsearch,liu2024eoh,ye2024reevo,vanstein2025llamea,yao2025moeoh,liu2026eohs}. In contrast, evolutionary PINN design has largely relied on predefined representations together with conventional mutation and crossover operators \citep{kaplarevic2023optimal,zhangyang2024evo,carrillo2024mopinns}. This suggests a natural opportunity to combine LLM-based semantic reasoning with evidence from real PDE solving to evolve complete PINN configurations across generations.

We therefore propose PINNForge, an execution-grounded evolutionary framework for automated PINN design. It starts from diverse configurations informed by PDE characteristics and relevant PINN knowledge, then executes them through real training to obtain performance, physical, and optimization evidence. PINNForge feeds globally strong configurations, their execution evidence, and accumulated run-level experience back to the LLM, which uses five complementary strategies---Refine, Physics-Guided, Architecture-Guided, Synthesis, and Novelty---to inherit, revise, recombine, or explore coupled PINN components. In this way, candidate training becomes part of the subsequent design process rather than a terminal evaluation. The main contributions are as follows:
\begin{enumerate}
    \item We formulate automated PINN design as an execution-grounded multi-generation search over a unified design space spanning representation, architecture, sampling, constraints, loss construction, optimization, and training control.
    
    \item We develop an EoH-inspired LLM-driven evolutionary mechanism in which PDE-related prior knowledge, globally retained high-performing configurations, and execution evidence guide cross-component inheritance, revision, recombination, and exploration.
    
    \item Across 25 heterogeneous PDEs, PINNForge achieves the lowest mean MSE on 24 tasks compared with RoPINN, PINNsFormer, and PINNsAgent. Controlled ablations, search trajectories, and training-budget analysis further examine the roles of its main search mechanisms.
\end{enumerate}

\section{Related Work}\label{sec:related_work}

\subsection{PINN Optimization and Automated Design}

PINN training can be hindered by gradient imbalance, mismatched convergence rates across loss terms, insufficient sampling, spectral bias, inaccurate constraint enforcement, and ill-conditioned optimization \citep{wang2021gradient,wang2022ntk,wu2023sampling,krishnapriyan2021failure,rathore2024challenges}. Prior work addresses these challenges through adaptive activations, Fourier features, adaptive loss weighting, exact boundary constraints, gradient-enhanced objectives, causal training, and domain-decomposition or variational formulations \citep{jagtap2020adaptive,wang2021fourier,mcclenny2023selfadaptive,sukumar2022exact,yu2022gradient,wang2024causal,jagtap2020xpinns,kharazmi2021hpvpinns}. These advances highlight the multi-component nature of effective PINN design, but individual methods typically target particular training difficulties specified in advance.

Automated approaches instead search for suitable PINN designs. Auto-PINN and NAS-PINN automate architecture and hyperparameter selection \citep{wang2023autopinn,wangzhong2024naspinn}, while evolutionary methods optimize architectures, activation functions, loss weights, and related hyperparameters \citep{kaplarevic2023optimal,zhangyang2024evo,carrillo2024mopinns}. More recently, PINNsAgent uses LLM-based knowledge replay and memory-guided reasoning to search PINN configurations, whereas Lang-PINN employs multiple agents for PDE understanding, code generation, execution, and correction \citep{wuwu2025pinnsagent,he2026langpinn}. PINNForge differs by treating execution evidence from real training as a persistent cross-generation design signal and by evolving complete, coupled PINN configurations through semantic inheritance, revision, recombination, and exploration rather than predefined parameter-level operators.

\subsection{LLM-Guided Evolutionary Algorithm Discovery}

LLMs are increasingly used within iterative algorithm-discovery systems. FunSearch repeatedly evaluates generated programs and reuses strong solutions as context \citep{romeraparedes2024funsearch}. EoH integrates LLM-based generation with evolutionary search in a generate--execute--evaluate--evolve loop \citep{liu2024eoh,yao2025moeoh,liu2026eohs}. ReEvo translates performance differences into language-based reflective feedback \citep{ye2024reevo}, while LLaMEA couples runtime evaluation, selection, and iterative modification for metaheuristic design \citep{vanstein2025llamea,vanstein2025intheloop}. Together, these studies show that LLMs can support semantic evolutionary operations by deciding which mechanisms to preserve, revise, combine, or replace based on candidate structure and execution behavior. PINNForge extends this principle to PINN configuration search, where each execution yields not only a scalar performance measure but also physically meaningful residual, constraint, and optimization evidence.

\section{Method}\label{sec:method}

\subsection{Overview}\label{sec:overview}

PINNForge adapts the generate--execute--evaluate--evolve principle of Evolution of Heuristics (EoH) to automated PINN design. Rather than generating a configuration once, it repeatedly trains candidate designs on the target PDE and uses the resulting performance, physical, and optimization evidence to guide subsequent generations. The search is therefore grounded in observed PDE-solving behavior rather than solely in static PDE descriptions or predefined parameter-level perturbations.

\begin{figure*}[t]
    \centering
    \includegraphics[width=\textwidth]{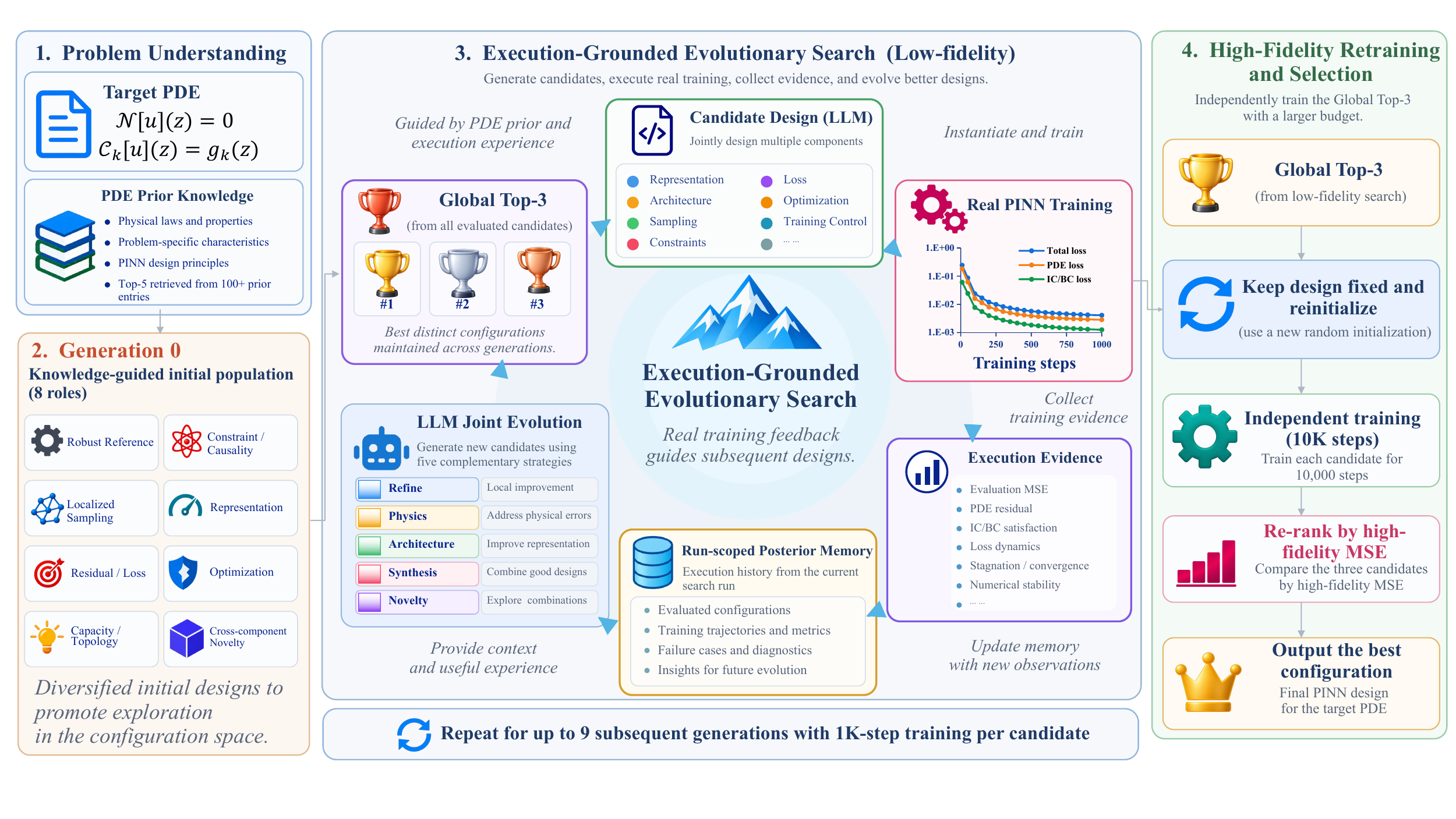}
    \caption{Overview of the PINNForge framework.}
    \label{fig:pinnsforge-framework}
\end{figure*}

Given a target PDE, PINNForge combines problem characteristics with relevant PINN prior knowledge to generate a diverse initial population. Each candidate is trained to obtain a scalar performance measure together with diagnostic execution evidence. The best distinct configurations over the full search history form the Global Top-3 and serve as parents for subsequent generations. The LLM then uses these elites, their execution evidence, and accumulated run-level experience to refine, recombine, or explore beyond existing designs. After low-fidelity search terminates, the final Global Top-3 configurations are independently retrained under a larger budget, and the best-performing high-fidelity candidate is returned. Figure~\ref{fig:pinnsforge-framework} summarizes the overall framework.

\subsection{Execution-Grounded Search Context}\label{sec:execution_context}

Let \(z\) denote the input coordinates and \(u(z)\in\mathbb{R}^{d_u}\) denote the physical field governed by a target PDE \(\mathcal{P}\). We express the governing equation and associated physical constraints as
\[
\mathcal{N}[u](z)=0,\qquad z\in\mathcal{D},
\]
and
\[
\mathcal{C}_k[u](z)=g_k(z),\qquad z\in\Gamma_k,
\]
respectively. A PINN \(u_\theta(z)\) approximates the physical field by minimizing
\[
\mathcal{L}(\theta)
=
\lambda_r\mathcal{L}_r
+
\sum_k \lambda_k\mathcal{L}_k,
\]
where \(\mathcal{L}_r\) denotes the PDE residual loss and \(\mathcal{L}_k\) denotes the loss associated with the \(k\)-th initial, boundary, or other physical constraint. Candidate accuracy is evaluated by mean squared error (MSE) on fixed evaluation points.

PINNForge conditions candidate generation on two complementary sources of information. The first is PDE-related prior knowledge. PINNForge maintains a fixed prior-knowledge base \(\mathcal{K}\) containing more than one hundred entries on general physical reasoning and PINN design principles. For a target PDE \(\mathcal{P}\), the method identifies properties relevant to PINN design---including time dependence, periodicity, multiscale structure, geometric complexity, locally steep behavior, and physical-constraint types---and retrieves the five most relevant entries from \(\mathcal{K}\). These entries form the PDE-specific prior context \(K_{\mathcal{P}}\), which provides design rationale rather than prescribing a fixed solver configuration and remains available throughout the search.

The second source is execution evidence obtained from candidates trained on the current PDE. For a candidate configuration \(a_i\), we denote its execution as
\[
\operatorname{Execute}(a_i;\mathcal{P})
\rightarrow
(y_i,E_i),
\]
where \(y_i\) is the scalar comparison metric used for ranking, and \(E_i\) records the physical and optimization diagnostics observed during training. Across candidates, these records capture PDE residuals, constraint satisfaction, convergence behavior, and numerical stability, providing diagnostic information about why a design succeeds or fails.

PINNForge uses execution evidence at two granularities. Selected parents retain candidate-specific evidence \(E_{P_{g+1}}\), while the full history through Generation \(g\) is compressed into a run-scoped posterior memory \(M_g\). The former preserves detailed evidence associated with specific parent configurations, whereas the latter summarizes recurring successes, failures, and design experience across the current search run. Appendix~A.3 details the evidence fields and the construction of \(M_g\).

\subsection{LLM-Driven Evolutionary PINN Design}\label{sec:evolutionary_design}

Each candidate is represented as a structured \texttt{AlgorithmSpec} spanning representation, architecture, sampling, physical constraints, loss construction, optimization, and training control. The LLM proposes specifications rather than arbitrary executable code, while a deterministic controller validates and maps compatible, budget-compliant specifications to supported PINN implementations. Appendix~A.1 summarizes the executable design dimensions and representative choices.

At Generation 0, no execution evidence from the current PDE is available. PINNForge therefore generates eight complete initial configurations under complementary design roles that emphasize different potential bottlenecks without prescribing fixed templates. Appendix~A.4 defines these roles.

After candidate execution, PINNForge performs global elite selection over the full search history. Let \(\mathcal{H}_g\) denote all candidates evaluated through Generation \(g\). The parent set for the next generation is
\[
P_{g+1}
=
\operatorname{GlobalDistinctTop3}
\left(
\mathcal{H}_g
\right).
\]
Selecting the three best distinct configurations over the full search history preserves strong early designs and provides multiple validated parents for subsequent refinement and recombination.

From Generation 1 onward, new candidates are conditioned on the Global Top-3, their candidate-specific execution evidence, PDE-related prior knowledge, and the run-scoped posterior memory. Because PINN components are coupled, PINNForge may revise several related components jointly rather than relying on independent numerical perturbations.

Each generation employs five complementary strategies. Refine makes conservative adjustments around the strongest design; Physics-Guided targets sampling, constraints, and loss construction; Architecture-Guided focuses on representation, architecture, and optimization; Synthesis combines complementary mechanisms from multiple strong parents; and Novelty explores beyond the current elite trajectory. Appendix~A.4 provides the full definitions.

For strategy \(s\), a next-generation candidate is generated as
\[
a_{g+1}^{(s)}
=
\operatorname{LLMGenerate}
\left(
P_{g+1},
E_{P_{g+1}},
K_{\mathcal{P}},
M_g,
s
\right),
\]
where \(E_{P_{g+1}}\) denotes the selected parents' candidate-specific execution evidence, \(K_{\mathcal{P}}\) the PDE-related prior knowledge, and \(M_g\) the run-scoped posterior memory summarized from the search history through Generation \(g\).

The strategy specifies the search direction, while the LLM decides which mechanisms to inherit, revise, recombine, or explore based on observed execution behavior. Executing the new candidates expands the history from \(\mathcal{H}_g\) to \(\mathcal{H}_{g+1}\), after which the posterior-summary step updates \(M_g\) to \(M_{g+1}\). The new candidates also compete with all previously evaluated candidates for the Global Top-3, closing the generate--execute--evaluate--evolve loop.

\subsection{Multi-Fidelity Search and High-Fidelity Evaluation}\label{sec:multifidelity}

Evaluating many candidates at the full training budget would be expensive. Inspired by multi-fidelity neural architecture search \citep{phan2024mfnas,won2025multifidelity}, PINNForge separates low-fidelity exploration from high-fidelity evaluation.

During low-fidelity search, every valid candidate receives the same \(1{,}000\)-step training budget, enabling efficient comparison while still producing execution evidence. Generation 0 contains eight candidates, followed by at most nine subsequent generations with five candidates each. The Global Top-3 is updated over the full search history.

Low-budget rankings may not perfectly predict fully trained performance \citep{yu2020nas}. PINNForge therefore does not directly return the best low-fidelity candidate. Once low-fidelity search terminates, the final Global Top-3 configurations are kept fixed, independently reinitialized, and retrained from scratch for \(10{,}000\) optimizer steps each. The candidate with the lowest high-fidelity evaluation MSE on the fixed reference points is then selected.

This two-stage procedure uses low-fidelity training for broad exploration and independent high-fidelity retraining to reduce dependence on transient low-budget rankings and particular parameter initializations.

\section{Experiments}\label{sec:experiments}

We evaluate PINNForge on 25 heterogeneous PDE benchmarks, focusing on solution accuracy, the roles of its main search mechanisms, and performance evolution across generations.

\subsection{Experimental Setup}\label{sec:experimental_setup}

Of the 25 PDEs, 21 are drawn from PINNacle \citep{hao2024pinnacle}, spanning nonlinear transport, reaction--diffusion, multiscale and long-time dynamics, complex geometries, high-dimensional PDEs, inverse problems, and fluid systems. We additionally include Allen--Cahn, variable-coefficient Darcy flow, the shallow-water system, and Kovasznay flow \citep{allen1979,darcy1856,saintvenant1871,kovasznay1948}. Definitions of all PDEs are provided in Appendix~D. We compare PINNForge with RoPINN, PINNsFormer, and PINNsAgent \citep{wu2024ropinn,zhao2024pinnsformer,wuwu2025pinnsagent}, representing PINN optimization, architectural design, and LLM-based automated configuration, respectively. Qwen3.8-27B is used as the fixed LLM backbone without PDE-specific fine-tuning.

For each PDE, solution accuracy is measured by MSE on fixed reference evaluation points. The same points are used for candidate comparison, search control, final configuration selection, and reporting; no separate held-out reference subset is introduced. We report mean MSE over 10 independent runs, each corresponding to one complete execution of the method, including its internal search and selection procedure. PINNForge trains each low-fidelity candidate for \(1{,}000\) optimizer steps and independently retrains each configuration in the final Global Top-3 for \(10{,}000\) steps before selecting the candidate with the lowest high-fidelity MSE. Appendix~B.2--B.4 provides details on LLM sampling, the computational environment, and the evaluation protocol.

Table~\ref{tab:main_results} reports mean$\pm$standard-deviation MSE over 10 independent runs on all 25 heterogeneous PDEs, together with aggregate statistical results. Appendix~B.6 provides per-task ranks, the PINNacle oracle reference, and additional statistical details.

\begin{table}[t]
\centering
\caption{Mean $\pm$ standard deviation of MSE over 10 independent runs on 25 PDEs, computed on the fixed reference evaluation points. Lower is better.}
\label{tab:main_results}
\scriptsize
\setlength{\tabcolsep}{2.5pt}
\renewcommand{\arraystretch}{0.90}
\begin{tabular}{@{}lcccc@{}}
\toprule
PDE & RoPINN & PINNsFormer & PINNsAgent & PINNForge \\
\midrule
burgers\_1d
& 7.31E-04$\pm$6.13E-05
& 1.08E-03$\pm$8.21E-06
& 6.52E-05$\pm$1.55E-05
& \textbf{6.50E-05$\pm$4.42E-07} \\

burgers\_2d
& 2.40E-01$\pm$3.12E-02
& 2.65E-01$\pm$6.83E-02
& 1.98E-01$\pm$1.66E-02
& \textbf{1.95E-01$\pm$1.31E-02} \\

gray\_scott\_2d
& 8.64E-03$\pm$4.53E-05
& 1.57E-02$\pm$1.32E-03
& 4.36E-03$\pm$3.08E-05
& \textbf{3.60E-03$\pm$2.65E-04} \\

heat\_2d\_complex\_geometry
& \textbf{4.00E-04$\pm$9.10E-05}
& 2.78E-03$\pm$5.77E-05
& 1.84E-03$\pm$9.95E-04
& 4.45E-04$\pm$9.72E-05 \\

heat\_2d\_long\_time
& 2.56E+00$\pm$4.53E-01
& 3.83E+00$\pm$1.64E-01
& 4.32E+00$\pm$2.78E+00
& \textbf{1.13E+00$\pm$4.21E-02} \\

heat\_2d\_multiscale
& 2.75E-05$\pm$7.52E-06
& 5.57E-05$\pm$9.93E-06
& 3.51E-05$\pm$2.20E-05
& \textbf{1.19E-05$\pm$2.16E-06} \\

heat\_2d\_varying\_coefficient
& 6.24E-01$\pm$4.53E-03
& 1.19E-02$\pm$3.42E-03
& 5.34E-03$\pm$3.85E-03
& \textbf{1.59E-03$\pm$4.77E-04} \\

heat\_5d
& 5.39E-07$\pm$1.31E-07
& 1.87E-07$\pm$1.91E-08
& 3.34E-07$\pm$7.65E-08
& \textbf{3.21E-08$\pm$3.26E-08} \\

heat\_inverse\_2d
& 4.31E-02$\pm$7.31E-03
& 8.89E-02$\pm$4.22E-02
& 6.32E-02$\pm$8.63E-02
& \textbf{7.15E-03$\pm$4.91E-03} \\

kuramoto\_sivashinsky\_1d
& 6.54E+00$\pm$5.14E-01
& 5.13E+00$\pm$4.51E-03
& 1.09E+00$\pm$3.18E-02
& \textbf{1.03E+00$\pm$8.40E-02} \\

navier\_stokes\_2d\_backstep
& 9.64E-03$\pm$5.13E-03
& 6.02E-04$\pm$4.33E-05
& 4.96E-03$\pm$1.45E-03
& \textbf{2.63E-04$\pm$4.71E-05} \\

navier\_stokes\_2d\_classic
& 2.88E-05$\pm$3.61E-06
& 1.49E-05$\pm$1.26E-05
& 8.48E-06$\pm$6.90E-06
& \textbf{2.48E-06$\pm$2.31E-06} \\

navier\_stokes\_2d\_long\_time
& 5.43E+02$\pm$5.63E+00
& 7.43E+02$\pm$1.31E+02
& 5.75E+02$\pm$7.63E+00
& \textbf{5.04E+02$\pm$4.95E+00} \\

poisson\_2d\_classic
& 8.57E-02$\pm$6.74E-02
& 2.17E-01$\pm$7.65E-02
& 6.34E-01$\pm$1.43E-01
& \textbf{1.61E-02$\pm$3.95E-04} \\

poisson\_boltzmann\_2d
& 7.91E-01$\pm$7.14E-01
& 1.34E-02$\pm$4.31E-03
& 6.97E-02$\pm$7.43E-02
& \textbf{5.38E-03$\pm$9.13E-04} \\

poisson\_2d\_many\_area
& 2.89E+00$\pm$3.51E-01
& 5.36E+00$\pm$3.79E+00
& 3.00E+00$\pm$1.01E+00
& \textbf{2.21E+00$\pm$3.56E-01} \\

poisson\_3d\_complex\_geometry
& 2.16E-01$\pm$1.53E-02
& 8.16E-03$\pm$5.58E-04
& 1.60E-02$\pm$1.08E-02
& \textbf{1.04E-03$\pm$2.34E-04} \\

poisson\_5d
& 4.09E-07$\pm$7.43E-08
& 2.92E-07$\pm$1.99E-07
& 2.01E-06$\pm$1.09E-06
& \textbf{5.64E-08$\pm$3.95E-08} \\

poisson\_inverse\_2d
& 5.32E-03$\pm$3.46E-03
& 1.51E-03$\pm$8.65E-04
& 2.41E-02$\pm$4.73E-02
& \textbf{2.57E-04$\pm$1.29E-04} \\

wave\_1d
& 4.19E-02$\pm$4.38E-03
& 3.61E-02$\pm$6.65E-03
& 3.25E-02$\pm$3.45E-02
& \textbf{3.18E-02$\pm$5.73E-03} \\

wave\_2d\_heterogeneous
& 4.63E-02$\pm$6.73E-03
& 8.37E-02$\pm$1.18E-02
& 5.17E-02$\pm$8.03E-03
& \textbf{2.84E-02$\pm$7.69E-04} \\

allen\_cahn\_1d
& 7.45E-08$\pm$4.61E-08
& 6.02E-09$\pm$1.08E-09
& 1.44E-08$\pm$3.50E-09
& \textbf{1.13E-09$\pm$2.26E-10} \\

darcy\_flow\_2d
& 3.48E-09$\pm$1.63E-09
& 1.35E-10$\pm$2.90E-11
& 5.33E-07$\pm$2.65E-07
& \textbf{1.64E-11$\pm$3.28E-12} \\

shallow\_water\_2d
& 5.78E-07$\pm$7.13E-07
& 2.24E-07$\pm$8.44E-08
& 6.81E-07$\pm$1.25E-07
& \textbf{7.15E-08$\pm$2.86E-08} \\

kovasznay\_flow\_2d
& 6.13E-07$\pm$4.31E-07
& 3.80E-08$\pm$9.63E-09
& 3.82E-07$\pm$3.66E-07
& \textbf{1.77E-09$\pm$4.43E-10} \\

\midrule
Friedman Avg. Rank & 3.04 & 2.96 & 2.96 & \textbf{1.04} \\
Wilcoxon vs. PINNForge & $+$ & $+$ & $+$ & -- \\
Median CV & 0.177 & 0.184 & 0.497 & 0.182 \\
\bottomrule
\end{tabular}

\vspace{2pt}
\begin{minipage}{0.98\linewidth}
\scriptsize
The Friedman test gives $\chi_F^2=42.696$ and $p=2.86\times10^{-9}$.
For Wilcoxon tests, ``$+$'' indicates that PINNForge significantly outperforms
the corresponding baseline under paired signed-rank tests with Holm correction
($p_{\mathrm{adj}}=1.79\times10^{-7}$ for each comparison).
CV denotes the median coefficient of variation across the 25 PDEs.
\end{minipage}
\end{table}

PINNForge achieves the lowest mean MSE among the four compared methods on 24 of 25 PDEs; RoPINN is slightly better only on \texttt{heat\_2d\_complex\_geometry}. Relative to the strongest baseline on each task, PINNForge yields a median MSE reduction of \(59.85\%\), with a geometric-mean best-baseline-to-PINNForge MSE ratio of \(2.78\times\). It also ranks first on all four PDEs added beyond PINNacle, indicating that the gains are not confined to a single equation family within this benchmark.

As summarized in Table~\ref{tab:main_results}, across the 25 PDEs, PINNForge achieves an average Friedman rank of \(1.04\), compared with \(3.04\), \(2.96\), and \(2.96\) for RoPINN, PINNsFormer, and PINNsAgent, respectively. A Friedman test on per-task ranks computed from the 10-run mean MSE indicates significant overall differences (\(\chi_F^2=42.696\), \(p=2.86\times10^{-9}\)). Pairwise Wilcoxon signed-rank tests on the \(\log_{10}\)-transformed per-task mean MSE across the 25 tasks, followed by Holm correction, favor PINNForge over all three baselines (\(p_{\mathrm{adj}}=1.79\times10^{-7}\) for each comparison). Median coefficients of variation are \(0.177\), \(0.184\), \(0.497\), and \(0.182\) for RoPINN, PINNsFormer, PINNsAgent, and PINNForge, respectively.

The magnitude of improvement remains task-dependent. Relative to the strongest baseline, PINNForge achieves \(3.36\times\) lower MSE on \texttt{heat\_2d\_varying\_coefficient}, \(5.83\times\) on \texttt{heat\_5d}, \(7.85\times\) on \texttt{poisson\_3d\_complex\_geometry}, \(8.23\times\) on \texttt{darcy\_flow\_2d}, and \(21.47\times\) on \texttt{kovasznay\_flow\_2d}, whereas the margins are small on \texttt{burgers\_1d}, \texttt{burgers\_2d}, and \texttt{wave\_1d}. This pattern is consistent with greater benefits on tasks that require coordinated choices across multiple PINN components, although these task-wise comparisons do not isolate which problem properties drive the gains.

\subsection{Component Ablation}\label{sec:ablation}

We evaluate three ablations on 12 representative PDEs to examine the roles of knowledge guidance, execution feedback, and parent-driven evolution. Knowledge guidance comprises the PDE prior \(K_{\mathcal{P}}\) and run-scoped posterior memory \(M_g\); \texttt{w/o Knowledge Guidance} removes both while retaining candidate-specific execution evidence and parent-driven evolution. \texttt{w/o Execution Feedback} withholds execution evidence from the LLM and consequently disables the execution-derived posterior memory, while preserving PDE-related prior knowledge and parent-driven evolution. \texttt{w/o Evolutionary Search} keeps the candidate-evaluation budget fixed but replaces parent-driven evolution with independent LLM generation. Table~\ref{tab:ablation_summary} summarizes these interventions and their aggregate effects. Appendix~A.6 details the information available to each variant, and Appendix~B.7 reports the complete per-task results.

\begin{table}[t]
\centering
\caption{Ablation summary over 12 representative PDEs. The last column reports the median task-wise MSE ratio relative to the full PINNForge; lower is better.}
\label{tab:ablation_summary}
\footnotesize
\setlength{\tabcolsep}{3.6pt}
\renewcommand{\arraystretch}{0.96}
\begin{tabular}{@{}lcccc@{}}
\toprule
Variant & Knowl. guidance & Exec. feedback & Parent evolution & Median ratio \\
\midrule
w/o Knowledge Guidance & $\times$ & $\checkmark$ & $\checkmark$ & $3.74\times$ \\
w/o Execution Feedback to LLM & $\checkmark$ & $\times$ & $\checkmark$ & $12.10\times$ \\
w/o Evolutionary Search & $\checkmark$ & $\checkmark$ & $\times$ & $10.10\times$ \\
PINNForge & $\checkmark$ & $\checkmark$ & $\checkmark$ & $1.00\times$ \\
\bottomrule
\end{tabular}
\end{table}

Removing Execution Feedback produces the largest aggregate degradation, increasing the median MSE to \(12.10\times\) that of full PINNForge. Because this variant withholds candidate-specific execution evidence and therefore cannot form the execution-derived posterior memory, the remaining PDE-related prior knowledge and parent structures are insufficient to recover full-method performance under this setup. Removing Evolutionary Search increases the median error to \(10.10\times\) under the same candidate-evaluation budget, consistent with a benefit from parent-driven inheritance and recombination. Removing Knowledge Guidance increases the median error to \(3.74\times\); because both \(K_{\mathcal{P}}\) and \(M_g\) are removed, this result reflects the knowledge-guidance channel as a whole rather than static prior knowledge alone.

Per-task results reveal distinct degradation patterns. On \texttt{heat\_2d\_complex\_geometry}, removing Execution Feedback increases MSE from \(4.45\times10^{-4}\) to \(1.59\), an increase of roughly \(3.6\times10^3\). Removing parent-driven evolution produces particularly large increases on \texttt{poisson\_5d} and \texttt{darcy\_flow\_2d}, where MSE rises by about \(365\times\) and \(652\times\), respectively. These cases are consistent with complementary roles for diagnostic feedback and inheritance, but do not constitute isolated causal attributions because multiple design components can change simultaneously.

\subsection{Training Budget}\label{sec:training_budget}

Table~\ref{tab:budget_summary} compares the optimizer-step budget required to obtain the selected result for a single PDE. Each of the three baselines trains five independently initialized candidates for \(20{,}000\) optimizer steps and selects the one with the lowest evaluation MSE, resulting in a total budget of \(100{,}000\) steps. PINNForge instead uses at most \(53{,}000\) steps for low-fidelity search and \(30{,}000\) steps for independent high-fidelity retraining of the final Global Top-3, yielding a maximum total of \(83{,}000\) optimizer steps.

\begin{table}[t]
\centering
\caption{PINN optimizer-step budget used to obtain the selected result for one PDE.}
\label{tab:budget_summary}
\footnotesize
\setlength{\tabcolsep}{4.0pt}
\renewcommand{\arraystretch}{0.96}
\begin{tabular}{@{}lll@{}}
\toprule
Method & Selection protocol & Budget \\
\midrule
RoPINN & Best of 5 independently initialized 20K-step candidates & 100K \\
PINNsFormer & Best of 5 independently initialized 20K-step candidates & 100K \\
PINNsAgent & Best of 5 independently initialized 20K-step candidates & 100K \\
PINNForge & 53$\times$1K LF + Top-3$\times$10K HF & $\leq$83K \\
\bottomrule
\end{tabular}
\end{table}

The reported gains therefore do not rely on a larger PINN optimizer-step budget than the baselines. This accounting compares optimization steps rather than total computational cost, since per-step cost can vary with network architecture, sampling scale, and optimizer configuration. It also excludes LLM inference, wall-clock time, and monetary cost; Appendix~B.8 provides the full accounting.

\subsection{Search Process Analysis}\label{sec:search_process}

\begin{figure}[t]
    \centering
    \includegraphics[width=\linewidth]{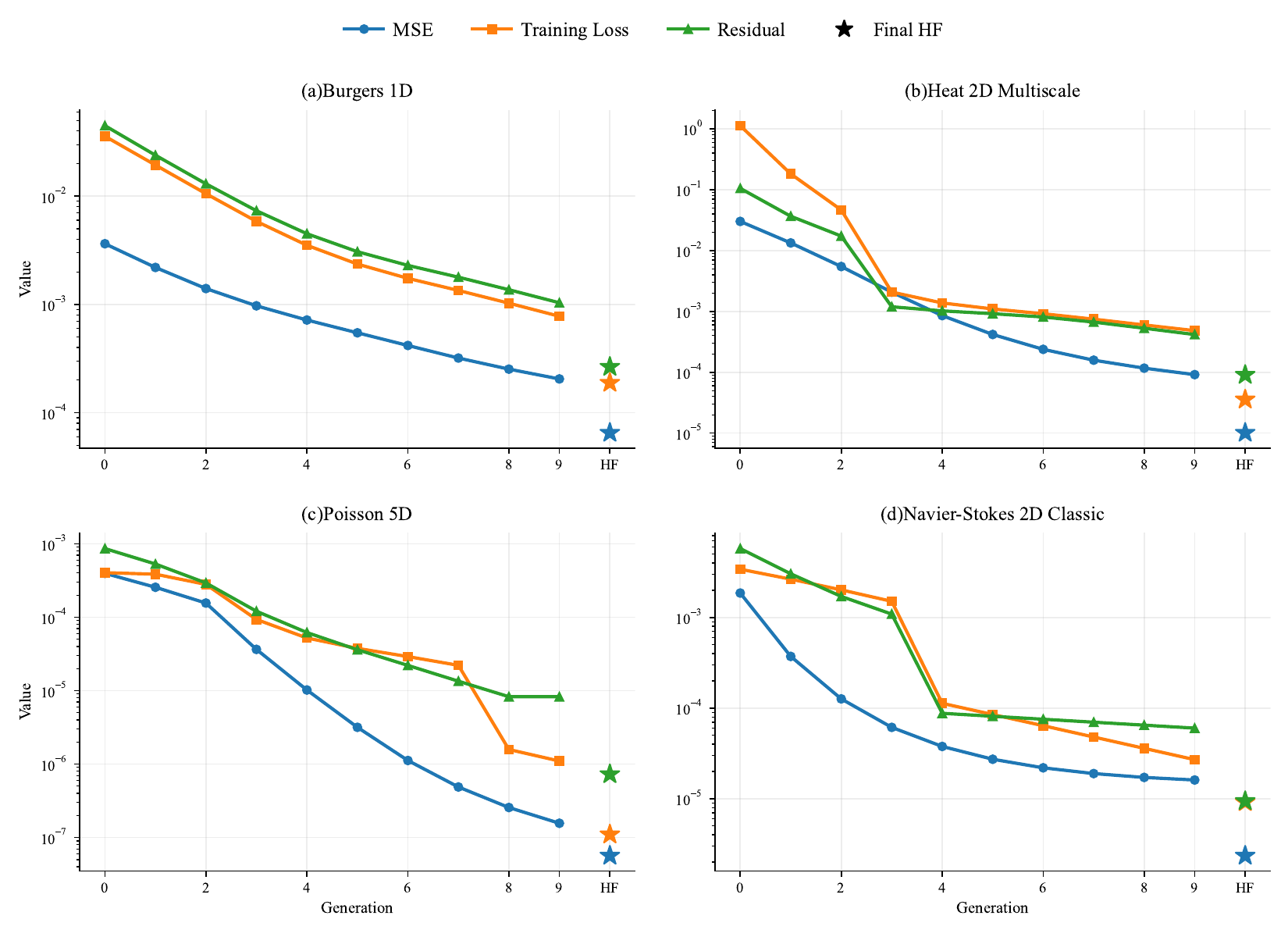}
    \caption{Search trajectories on four representative PDEs. At each generation, the candidate with the lowest low-fidelity MSE is selected, and its MSE, training loss, and PDE residual are shown. Stars mark the corresponding results after independent high-fidelity retraining of the final Global Top-3.}
    \label{fig:search-trajectories}
\end{figure}

We next examine whether performance continues to improve beyond Generation 0 on four representative PDEs spanning nonlinear transport, multiscale dynamics, high-dimensional PDEs, and fluid systems. These trajectories are obtained from separate single-run searches conducted specifically for process analysis and are independent of the ten-run experiments used for aggregate performance statistics. Figure~\ref{fig:search-trajectories} tracks the candidate with the lowest low-fidelity MSE in each generation, together with its corresponding training loss and PDE residual. Stars mark the corresponding results after independent high-fidelity retraining of the final Global Top-3. Appendix~C reports the corresponding trajectories for all 25 PDEs.

All four representative searches discover a new low-fidelity global best after Generation 0. The training loss and PDE residual associated with the generation-best MSE candidates also decrease broadly across the four searches, indicating that the reduction in evaluation error is generally accompanied by improvements in the training objective and PDE consistency. As summarized in Table~\ref{tab:trajectory_summary}, low-fidelity evolution reduces MSE by \(17.76\times\)--\(2{,}538.46\times\) relative to the best Generation-0 candidate, while high-fidelity retraining increases the total improvement to \(55.81\times\)--\(7{,}008.58\times\).

\begin{table}[t]
\centering
\caption{Representative single-run trajectories. G0 is the best Generation-0 evaluation MSE, LF the best low-fidelity MSE, and HF the best MSE after high-fidelity retraining.}
\label{tab:trajectory_summary}
\footnotesize
\setlength{\tabcolsep}{3.3pt}
\renewcommand{\arraystretch}{0.96}
\begin{tabular}{@{}lccccc@{}}
\toprule
PDE & G0 & Best LF & Final HF & G0$\rightarrow$HF & LF$\rightarrow$HF \\
\midrule
burgers\_1d & 3.64E-03 & 2.05E-04 & 6.52E-05 & $55.81\times$ & $3.15\times$ \\
heat\_2d\_multiscale & 3.02E-02 & 9.14E-05 & 1.01E-05 & $2{,}986.95\times$ & $9.05\times$ \\
poisson\_5d & 3.96E-04 & 1.56E-07 & 5.65E-08 & $7{,}008.58\times$ & $2.77\times$ \\
navier\_stokes\_2d\_classic & 1.86E-03 & 1.61E-05 & 2.35E-06 & $792.89\times$ & $6.85\times$ \\
\bottomrule
\end{tabular}
\end{table}

The Burgers-1D trajectory illustrates how execution evidence informs subsequent design decisions. For parent \texttt{G6-C4}, training exhibited stable optimization and strong hard-constraint satisfaction but a persistent residual hotspot near the shock. Child \texttt{G7-C4} therefore preserved the stable architecture and constraint mechanisms while strengthening local adaptive sampling and revising the optimization configuration; solution MSE decreased from \(5.71\times10^{-4}\) to \(2.21\times10^{-4}\), while the PDE residual decreased from \(6.05\times10^{-2}\) to \(2.14\times10^{-2}\). Because the child is independently initialized and modifies both sampling and optimization, this trace does not attribute the improvement to any single component. Appendix~A.7 provides the complete component-level trace and evolutionary lineage.

The additional \(2.77\times\)--\(9.05\times\) improvement from Best LF to Final HF shows that configurations identified under the low-fidelity budget can improve further under longer independent optimization. Together with the generation-wise gains, these trajectories provide evidence that performance improvements continue beyond the initial population, while not isolating the causal contribution of any individual search mechanism.

\subsection{Discussion}\label{sec:discussion}

These analyses provide complementary evidence. The 25-PDE benchmark establishes the overall accuracy gains, the budget-matched ablations show that candidate-evaluation budget alone does not account for the advantage, the search trajectories demonstrate continued improvement beyond Generation 0, and the Burgers trace illustrates how execution evidence informs subsequent design decisions. Together, these results are consistent with a beneficial interaction between execution evidence and parent-driven evolution over historical elites, although they do not isolate the causal contribution of individual mechanisms.

The gains remain task-dependent, with RoPINN slightly outperforming PINNForge on \texttt{heat\_2d\_complex\_geometry}. Because the same fixed reference points are reused for search control, final selection, and reporting, the reported MSE reflects performance under this benchmark protocol rather than held-out generalization, and adaptive reuse may favor configurations tuned to these points. The study also uses a single LLM backbone, Qwen3.8-27B, and compares PINN optimizer-step budgets rather than end-to-end computational cost. Cross-backbone robustness, independently sampled evaluation points, and end-to-end efficiency therefore remain important directions for further evaluation.

\section{Conclusion}\label{sec:conclusion}

We developed PINNForge, an execution-grounded evolutionary framework for automated PINN design that integrates PDE-related prior knowledge, historical elites, execution evidence, and LLM-driven cross-component evolution. Rather than treating candidate training as a terminal evaluation step, PINNForge feeds observed performance, physical, and optimization evidence into subsequent generations, enabling complete PINN configurations to be iteratively revised, recombined, and explored within a closed loop.

Across 25 heterogeneous PDEs, PINNForge achieves the lowest mean MSE on 24 tasks with an average Friedman rank of \(1.04\), while using at most \(83{,}000\) PINN optimizer steps per run. Controlled ablations show substantial degradation when knowledge guidance, execution feedback, or parent-driven evolution is removed, while search trajectories demonstrate that strong configurations continue to emerge beyond Generation~0 and can improve further under independent high-fidelity retraining. Together, these results support execution-grounded cross-generation evolution as a promising direction for moving automated PINN design beyond one-shot configuration generation toward iterative design improvement driven by observed PDE-solving behavior.

\clearpage
\bibliography{references}
\bibliographystyle{plainnat}

\end{document}


\maketitle
\appendix
\section{Method Implementation and Search Mechanics}\label{appendix-a-method-implementation-and-search-mechanics}

This appendix specifies the executable candidate representation, LLM context, role-conditioned search directions, controller-side validation, ablation isolation, and a representative cross-generation trace. It also separates decisions made by the LLM from those fixed by the deterministic controller.

\subsection{\texttt{AlgorithmSpec} and Executable Design Dimensions}\label{a.1-algorithmspec-search-space}

PINNForge represents each candidate as a structured \texttt{AlgorithmSpec} rather than arbitrary executable code. The LLM proposes controller-supported components, and a candidate is executed only after deterministic normalization, compatibility checks, and budget validation. We denote the controller-supported configuration space by \(\Omega\).

The \texttt{AlgorithmSpec} jointly covers the main PINN design dimensions. The table below lists representative executable choices exposed by the implementation; these entries describe each dimension's scope rather than a fixed PDE-specific template.

\begin{longtable}[]{@{}
  >{\raggedright\arraybackslash}p{0.24\columnwidth}
  >{\raggedright\arraybackslash}p{0.70\columnwidth}@{}}
\toprule\noalign{}
Design dimension & Representative executable choices \\
\midrule\noalign{}
\endhead
\bottomrule\noalign{}
\endlastfoot
Representation & Coordinate transforms, Fourier or periodic features, and multiscale representations supported by the implementation. \\
Architecture & Network depth and width, activation functions, residual connections, and other supported topology-related choices. \\
Sampling & Collocation-point allocation, sampling distributions, and adaptive or localized resampling mechanisms. \\
Constraints & Soft constraints, exact constraint transforms when available, and supported periodic or interface handling. \\
Loss & PDE-residual construction, loss weighting, normalization, causal weighting, and other supported physical-loss handling. \\
Optimization & Optimizer choice, learning rate, gradient handling, scheduling, and staged optimization. \\
Training control & Resampling, curriculum or stage allocation, checkpoint-related control, and other supported training procedures. \\
\end{longtable}

These dimensions are searched jointly. A proposal may preserve some components while modifying several coupled components at once---for example, representation, sampling, and optimization---when the observed training behavior suggests that a scalar adjustment is insufficient. Every candidate remains restricted to controller-supported implementations in \(\Omega\).

The controller also enforces resource limits that the LLM cannot relax: at most 2M trainable parameters, 65,536 training points, and fixed low- and high-fidelity budgets of 1K and 10K optimizer steps, respectively (Appendix~\ref{appendix-b-experimental-protocol-results}). Thus, \(\Omega\) defines the admissible design space, while the experimental protocol defines its resource envelope.

\subsection{LLM Inputs, Outputs, and Information Boundaries}\label{a.2-llm-inputs-outputs-boundaries}

Each LLM call proposes one complete structured \texttt{AlgorithmSpec}; it does not generate arbitrary executable code. A controller handles parsing, validation, instantiation, and real PINN training, preventing the LLM from bypassing the search space or modifying the experimental protocol.

The information provided to the LLM at different search stages is summarized below.

\begin{longtable}[]{@{}
  >{\raggedright\arraybackslash}p{0.22\columnwidth}
  >{\centering\arraybackslash}p{0.12\columnwidth}
  >{\centering\arraybackslash}p{0.14\columnwidth}
  >{\raggedright\arraybackslash}p{0.42\columnwidth}@{}}
\toprule\noalign{}
\begin{minipage}[b]{\linewidth}\raggedright
Context
\end{minipage} & \begin{minipage}[b]{\linewidth}\centering
Generation 0
\end{minipage} & \begin{minipage}[b]{\linewidth}\centering
Generation \(g\ge 1\)
\end{minipage} & \begin{minipage}[b]{\linewidth}\raggedright
Purpose
\end{minipage} \\
\midrule\noalign{}
\endhead
\bottomrule\noalign{}
\endlastfoot
Target PDE description & \(\checkmark\) & \(\checkmark\) & Provides the governing equation, computational domain, IC/BC, and problem characteristics \\
Prior knowledge \(K_{\mathcal{P}}\) & \(\checkmark\) & \(\checkmark\) & Provides PINN design rationale relevant to the current PDE \\
Allowed search space \(\Omega\) & \(\checkmark\) & \(\checkmark\) & Restricts the network, sampling, constraint, loss, and optimization components that may be generated \\
Training / resource budget & \(\checkmark\) & \(\checkmark\) & Restricts parameter count, sampling scale, and training budget \\
Role / search-strategy condition & \(\checkmark\) & \(\checkmark\) & Specifies the design focus of the current candidate \\
Global Top-3 \texttt{AlgorithmSpec} & -- & \(\checkmark\) & Provides validated high-performing parent configurations \\
Candidate-specific execution evidence of parents & -- & \(\checkmark\) & Provides raw parent-linked diagnostics including PDE residuals, IC/BC errors, optimization stability, and convergence behavior \\
Run-scoped posterior memory \(M_g\) & -- & \(\checkmark\) & Provides a compact cross-candidate summary of successes, failures, and design experience observed within the current independent search run \\
\end{longtable}

Generation 0 has no execution history, so initial candidates use only PDE information, prior knowledge, the allowed search space, resource constraints, and initialization roles. From Generation 1 onward, the context also includes the Global Top-3 parents, their execution evidence, and the run-scoped posterior memory. Appendix~\ref{a.4-initialization-roles-evolution-strategies} defines the roles and evolutionary strategies; Appendix~\ref{b.2-llm-sampling} gives their sampling temperatures.

The LLM cannot modify the target PDE, domain, initial or boundary conditions, evaluation protocol, or experimental budget. Its output is restricted to PINN components in \(\Omega\), including representation, architecture, activation, constraints, sampling, loss handling, optimization, and training control. Candidates are normalized before deduplication; exactly identical normalized \texttt{AlgorithmSpec}s are treated as one configuration, avoiding duplicates caused only by field ordering or expanded defaults.

\subsection{PDE-Related Prior Knowledge and Run-Scoped Execution Memory}\label{a.3-prior-knowledge-run-scoped-memory}

PINNForge separates information available before execution from empirical evidence collected during the current search run. This distinction keeps general design rationale separate from problem-specific training experience.

\paragraph{PDE-related prior knowledge.}
For a target PDE \(\mathcal{P}\), PINNForge characterizes properties relevant to PINN design, including time dependence, periodicity, multiscale structure, geometric complexity, locally steep behavior, and physical-constraint types. Associated design knowledge is supplied as \(K_{\mathcal{P}}\), which provides physical and numerical rationale for design hypotheses without prescribing a specific architecture, optimizer, sampling rule, or complete solver. The same prior remains available in later generations alongside execution evidence.

\paragraph{Candidate-level execution evidence.}
After a candidate is trained on \(\mathcal{P}\), execution returns \((y_i,E_i)\). Here, \(y_i\) is the scalar comparison metric used for ranking, while \(E_i\) records diagnostics such as PDE residual, constraint satisfaction, loss reduction, stagnation, gradient or loss imbalance when available, convergence behavior, and numerical stability. Keeping \(y_i\) separate from \(E_i\) distinguishes ranking from the diagnostic context used to guide later design changes.

\paragraph{Run-scoped posterior memory.}
Let \(\mathcal{H}_g\) be the complete history through Generation \(g\), including validated \texttt{AlgorithmSpec}s and execution results. A separate posterior-summary LLM call compresses recurring successes, failures, physical-constraint behavior, and optimization experience in \(\mathcal{H}_g\) into run-scoped memory \(M_g\). This call only summarizes evidence from the current independent run; it neither generates a candidate nor changes controller-side rankings. Unlike \(K_{\mathcal{P}}\), \(M_g\) contains no experience from other runs and is absent at Generation 0.

Candidate-specific evidence and posterior memory therefore serve different roles. For Generation \(g+1\), \(E_{P_{g+1}}\) provides detailed evidence for the selected Global Top-3 parents, whereas \(M_g\) summarizes the broader history \(\mathcal{H}_g\). After Generation \(g+1\) executes, the history becomes \(\mathcal{H}_{g+1}\) and the posterior-summary call produces \(M_{g+1}\). Thus, \(K_{\mathcal{P}}\), parent evidence, and \(M_g\) provide prior, local, and run-level context at distinct granularities. Appendix~\ref{b.4-evaluation-reporting-protocol} specifies the controller-defined evaluation protocol.

\subsection{Initialization Roles and Evolution Strategies}\label{a.4-initialization-roles-evolution-strategies}

\paragraph{Generation-0 initialization roles.}
Generation 0 uses eight complementary roles to broaden initial coverage. Each role sets the design emphasis of one LLM call rather than a fixed solver template. All roles receive the same PDE description, \(K_{\mathcal{P}}\), allowed design space, and resource constraints, and each produces one complete \texttt{AlgorithmSpec}.

\begin{longtable}[]{@{}
  >{\raggedright\arraybackslash}p{0.30\columnwidth}
  >{\raggedright\arraybackslash}p{0.64\columnwidth}@{}}
\toprule\noalign{}
Generation-0 role & Primary design emphasis \\
\midrule\noalign{}
\endhead
\bottomrule\noalign{}
\endlastfoot
Robust reference & Construct a conservative reference design that prioritizes broadly applicable mechanisms and stable training. \\
Constraint / causality & Emphasize physical-constraint enforcement and, when relevant, temporal or causal structure. \\
Localized sampling & Emphasize local structure, collocation-point placement, and adaptive or localized refinement of difficult regions. \\
Representation & Emphasize network input or feature representation and its interaction with the solver architecture. \\
Residual / loss & Emphasize PDE-residual construction, physical-loss formulation, and loss-handling choices. \\
Optimization & Emphasize optimizer selection, scheduling, gradient behavior, and numerical training stability. \\
Capacity / topology & Emphasize model capacity, network depth and width, and supported topology-related choices. \\
Cross-component novelty & Explore less conventional combinations spanning multiple executable PINN components. \\
\end{longtable}

The eight candidates form the knowledge-guided initial population for subsequent execution-grounded evolution.

\paragraph{Evolution strategies for \(g\geq 1\).}
After execution, PINNForge maintains the distinct Global Top-3 over the full search history. From Generation 1 onward, each new candidate is conditioned on these parents, their candidate-specific evidence, \(K_{\mathcal{P}}\), \(M_g\), and one of five evolutionary strategies. For strategy \(s\),
\[
a_{g+1}^{(s)}=
\operatorname{LLMGenerate}\!\left(
P_{g+1},\,E_{P_{g+1}},\,K_{\mathcal{P}},\,M_g,\,s
\right),
\]
where \(P_{g+1}\) is the Global Top-3 parent set, \(E_{P_{g+1}}\) is its candidate-specific execution evidence, and \(M_g\) summarizes the broader history \(\mathcal{H}_g\).

\begin{longtable}[]{@{}
  >{\raggedright\arraybackslash}p{0.26\columnwidth}
  >{\raggedright\arraybackslash}p{0.68\columnwidth}@{}}
\toprule\noalign{}
Evolution strategy & Primary evolution direction \\
\midrule\noalign{}
\endhead
\bottomrule\noalign{}
\endlastfoot
Refine & Use the current strongest candidate as the primary parent, preserve effective existing mechanisms, and make limited adjustments targeting its main shortcomings. \\
Physics-Guided & Adjust sampling-, constraint-, and loss-related designs primarily according to the observed PDE residual and initial/boundary-condition errors. \\
Architecture-Guided & Adjust network representation, architecture, and related training configurations according to representation capacity, model capacity, and optimization behavior. \\
Synthesis & Combine complementary, well-performing design mechanisms from multiple Global Top-3 parents into a new complete configuration. \\
Novelty & Explore design combinations not yet covered by the current high-performing candidates or the existing search trajectory. \\
\end{longtable}

The strategies define semantic search preferences rather than field-wise mutation or crossover operators. The LLM decides what to preserve, modify jointly, or recombine from multiple parents using PDE characteristics and observed execution behavior. After execution, new results extend the history to \(\mathcal{H}_{g+1}\), the posterior-summary call updates \(M_g\) to \(M_{g+1}\), and all candidates compete for the Global Top-3. This closes the generate--execute--evaluate--evolve loop.

The role- and strategy-specific sampling temperatures, including the more exploratory escape-generation schedule, are listed in Appendix~\ref{b.2-llm-sampling}.

\subsection{Deterministic Validation and Normalization}\label{a.5-validation-normalization}

The LLM only proposes candidates; a deterministic controller decides whether each proposal is admitted to training. Every returned \texttt{AlgorithmSpec} passes through

\[
\text{Parsing}
\rightarrow
\text{Normalization}
\rightarrow
\text{Validation}
\rightarrow
\text{Instantiation}
\rightarrow
\text{Training}.
\]

The validation stage checks at least the following conditions:

\begin{itemize}
\tightlist
\item
  whether the structure of the \texttt{AlgorithmSpec} is complete;
\item
  whether all components belong to the allowed search space \(\Omega\);
\item
  whether different algorithm components are mutually compatible;
\item
  whether the number of network parameters satisfies the budget;
\item
  whether the sampling budget satisfies the constraints;
\item
  whether the training iteration count strictly matches the current low- or high-fidelity training budget.
\end{itemize}

Only candidates that pass deterministic validation are instantiated and trained under the experimental protocol, producing \((y_{g,k},E_{g,k})\) for later search. LLM suggestions alone are not valid candidates: only \texttt{AlgorithmSpec}s that pass deterministic validation and complete real PINN training enter the history and Global Top-3.

\subsection{Information Isolation in Ablation Experiments}\label{a.6-ablation-information-isolation}

Section 4.2 defines three ablations. To focus each intervention on designated information channels, we keep the candidate count, training budget, search space, and LLM generation protocol unchanged while varying the context available to the LLM.

\textbf{Table A.1: Information accessible to the LLM under different ablation settings.}

\begin{longtable}[]{@{}
  >{\raggedright\arraybackslash}p{(\columnwidth - 8\tabcolsep) * \real{0.2190}}
  >{\centering\arraybackslash}p{(\columnwidth - 8\tabcolsep) * \real{0.1429}}
  >{\centering\arraybackslash}p{(\columnwidth - 8\tabcolsep) * \real{0.1905}}
  >{\centering\arraybackslash}p{(\columnwidth - 8\tabcolsep) * \real{0.2381}}
  >{\centering\arraybackslash}p{(\columnwidth - 8\tabcolsep) * \real{0.2095}}@{}}
\toprule\noalign{}
\begin{minipage}[b]{\linewidth}\raggedright
Variant
\end{minipage} & \begin{minipage}[b]{\linewidth}\centering
Prior Knowledge
\end{minipage} & \begin{minipage}[b]{\linewidth}\centering
Run-scoped Posterior
\end{minipage} & \begin{minipage}[b]{\linewidth}\centering
Execution Feedback to LLM
\end{minipage} & \begin{minipage}[b]{\linewidth}\centering
Parent-based Evolution
\end{minipage} \\
\midrule\noalign{}
\endhead
\bottomrule\noalign{}
\endlastfoot
w/o Knowledge Guidance & \(\times\) & \(\times\) & \(\checkmark\) & \(\checkmark\) \\
w/o Execution Feedback to LLM & \(\checkmark\) & \(\times\) & \(\times\) & \(\checkmark\) \\
w/o Evolutionary Search & \(\checkmark\) & \(\checkmark\) & \(\checkmark\) & \(\times\) \\
\textbf{PINNForge} & \(\checkmark\) & \(\checkmark\) & \(\checkmark\) & \(\checkmark\) \\
\end{longtable}

In \texttt{w/o\ Knowledge\ Guidance}, the LLM receives neither \(K_{\mathcal{P}}\) nor \(M_g\), while raw execution evidence and parent-based evolution are retained. This variant therefore evaluates the knowledge-guidance channel as a whole rather than isolating static prior knowledge from the execution-derived posterior.

In \texttt{w/o\ Execution\ Feedback}, historical candidate structures remain visible, but \(y_{g,k}\), \(E_{g,k}\), residuals, constraint errors, optimization diagnostics, rankings, and other training outcomes are hidden. Because \(M_g\) depends on this history, no posterior memory is formed. The LLM therefore retains prior knowledge and parent structures without access to their observed execution behavior.

In \texttt{w/o\ Evolutionary\ Search}, the LLM retains aggregated historical execution experience but receives no historical \texttt{AlgorithmSpec}s or Global Top-3 parents. Candidates are generated independently, without inheriting, modifying, or recombining existing high-performing configurations, under the same candidate-evaluation budget.

\subsection{Execution Feedback and Cross-Generation Inheritance Evidence on Burgers-1D}\label{a.7-burgers-mechanistic-evidence}

The main text gives a compact Burgers-1D parent--child example and shows that later generations continue to find lower-MSE candidates after Generation 0; Appendix~\ref{appendix-c-complete-search-trajectories} reports all 25 trajectories. Here we provide the full component-level trace and the final candidate's cross-generation lineage, showing how training feedback informs later design changes and how parent inheritance contributes to the final configuration.

Below, we use \texttt{Gg-Ck} as a readable label for candidate \(a_{g,k}\); for example, \texttt{G6-C4} corresponds to \(a_{6,4}\).

\paragraph*{\texorpdfstring{\textbf{A.7.1 Representative Parent---Child Modification}}{A.7.1 Representative Parent---Child Modification}}\label{a.7.1-parent-child-modification}

For parent \texttt{G6-C4}, the observed run shows numerically stable optimization and very small hard-constraint errors, but a clear PDE-residual hotspot remains near the shock. The child preserves the stable network and constraint mechanisms, strengthens local adaptive sampling, and explores an alternative optimization configuration.

\textbf{Table A.2: Representative execution-feedback-driven modification on Burgers-1D.}

\begin{longtable}[]{@{}
  >{\raggedright\arraybackslash}p{(\columnwidth - 8\tabcolsep) * \real{0.2000}}
  >{\raggedright\arraybackslash}p{(\columnwidth - 8\tabcolsep) * \real{0.2000}}
  >{\raggedright\arraybackslash}p{(\columnwidth - 8\tabcolsep) * \real{0.2000}}
  >{\raggedright\arraybackslash}p{(\columnwidth - 8\tabcolsep) * \real{0.2000}}
  >{\raggedright\arraybackslash}p{(\columnwidth - 8\tabcolsep) * \real{0.2000}}@{}}
\toprule\noalign{}
\begin{minipage}[b]{\linewidth}\raggedright
Component
\end{minipage} & \begin{minipage}[b]{\linewidth}\raggedright
Parent Configuration
\end{minipage} & \begin{minipage}[b]{\linewidth}\raggedright
Execution Evidence
\end{minipage} & \begin{minipage}[b]{\linewidth}\raggedright
Design Decision
\end{minipage} & \begin{minipage}[b]{\linewidth}\raggedright
Child Configuration
\end{minipage} \\
\midrule\noalign{}
\endhead
\bottomrule\noalign{}
\endlastfoot
Architecture & Modified MLP, \(6\times128\); no residual connections & Completed the full 1,000-step training without NaN or solution collapse & \textbf{Preserve.} The architecture was numerically stable in the observed run & Keep Modified MLP, \(6\times128\) \\
Constraint & Exact hard transform \texttt{burgers\_1d\_initial\_dirichlet\_exact} & IC error \(=2.25\times10^{-8}\), BC error \(=8.74\times10^{-8}\) & \textbf{Preserve.} Do not reallocate search emphasis to constraints that are already well satisfied & Keep the exact hard transform \\
Sampling & 10,000 Sobol interior points + RAD; add 500 each time; cap at 15,000; residual exponent 2.5 & Persistent residual hotspot at \(x\in[-0.24,0.24]\), \(t\in[0.15,0.39]\) & \textbf{Modify.} Strengthen local refinement while preserving global Sobol coverage & Add 600 each time; raise the cap to 16,000; keep the remaining RAD settings unchanged \\
Optimization & Adam 800 steps, lr \(1.0\times10^{-3}\), one-cycle; L-BFGS 200 steps, history 100 & Optimization remains numerically stable, but clear residual error remains at the end of training & \textbf{Explore.} Treat optimization changes as accompanying exploration rather than misclassifying a stable optimizer as a failure & Increase Adam lr to \(1.2\times10^{-3}\), switch to step decay; L-BFGS history 120 \\
\end{longtable}

The actual training results of the parent and child are:

\begin{longtable}[]{@{}lrr@{}}
\toprule\noalign{}
Metric & Parent G6-C4 & Child G7-C4 \\
\midrule\noalign{}
\endhead
\bottomrule\noalign{}
\endlastfoot
Solution MSE & \(5.71\times10^{-4}\) & \(2.21\times10^{-4}\) \\
PDE residual & \(6.05\times10^{-2}\) & \(2.14\times10^{-2}\) \\
IC error & \(2.25\times10^{-8}\) & \(1.25\times10^{-8}\) \\
BC error & \(8.74\times10^{-8}\) & \(2.53\times10^{-8}\) \\
\end{longtable}

This example does not attribute the improvement to a single component: parent and child use independent random initializations, and both sampling and optimization change. Instead, it records how execution feedback distinguishes mechanisms to preserve from components that remain candidates for modification.

\paragraph*{\texorpdfstring{\textbf{A.7.2 Cross-Generation Evolutionary Lineage of the Final Candidate}}{A.7.2 Cross-Generation Evolutionary Lineage of the Final Candidate}}\label{a.7.2-cross-generation-lineage}

PINNForge records the parent source of each evolutionary candidate. Single-parent refinement is written as

\[
a_{g,p}\rightarrow a_{g+1,c},
\]

Multi-parent synthesis inherits and recombines mechanisms from several Global Top-3 candidates. Refinement and synthesis are not predefined field-wise operators; the LLM decides what to preserve, modify, or combine from the parent \texttt{AlgorithmSpec}s, target PDE, and execution evidence.

\textbf{Table A.3: Cross-generation evolutionary lineage of the final Burgers-1D candidate.}

\begin{longtable}[]{@{}
  >{\raggedleft\arraybackslash}p{(\columnwidth - 8\tabcolsep) * \real{0.2500}}
  >{\raggedright\arraybackslash}p{(\columnwidth - 8\tabcolsep) * \real{0.1875}}
  >{\raggedright\arraybackslash}p{(\columnwidth - 8\tabcolsep) * \real{0.1875}}
  >{\raggedright\arraybackslash}p{(\columnwidth - 8\tabcolsep) * \real{0.1875}}
  >{\raggedright\arraybackslash}p{(\columnwidth - 8\tabcolsep) * \real{0.1875}}@{}}
\toprule\noalign{}
\begin{minipage}[b]{\linewidth}\raggedleft
Generation
\end{minipage} & \begin{minipage}[b]{\linewidth}\raggedright
Candidate
\end{minipage} & \begin{minipage}[b]{\linewidth}\raggedright
Parent(s)
\end{minipage} & \begin{minipage}[b]{\linewidth}\raggedright
Evolution Mode
\end{minipage} & \begin{minipage}[b]{\linewidth}\raggedright
Main Inherited / Modified Components
\end{minipage} \\
\midrule\noalign{}
\endhead
\bottomrule\noalign{}
\endlastfoot
0 & G0-C2 & -- & Independent Initialization & 4 $\times$ 64 modified MLP, \texttt{tanh}, soft constraints, Sobol + RAD, Adam + L-BFGS \\
1 & G1-C3 & G0-C2, G0-C6, G0-C4 & Multi-Parent Synthesis & Combine RAD sampling, a deeper network topology, and optimization control; expand to a 6 $\times$ 96 modified MLP \\
2 & G2-C1 & G1-C3 & Single-Parent Refinement & Preserve the 6 $\times$ 96 network; increase the RAD exponent and interior-point budget \\
3 & G3-C4 & G2-C1, G2-C4 & Multi-Parent Synthesis & Introduce an exact hard IC/BC transform; use fixed weighting; increase the RAD cap \\
4 & G4-C4 & G3-C4, G3-C2 & Multi-Parent Synthesis & Preserve hard constraints and Sobol + RAD; widen the network to 6 $\times$ 128; add one-cycle Adam and L-BFGS recovery \\
5 & G5-C4 & G4-C4, G4-C3 & Multi-Parent Synthesis & Adjust the Adam schedule, RAD capacity, and L-BFGS history \\
6 & G6-C4 & G5-C4, G5-C1, G5-C3 & Multi-Parent Synthesis & Preserve hard constraints, \texttt{tanh}, Sobol + RAD, and Adam$\rightarrow$L-BFGS; use a stable 6 $\times$ 128 modified MLP \\
7 & G7-C4 & G6-C4 & Single-Parent Refinement & Strengthen RAD refinement; adjust the Adam schedule and L-BFGS history \\
8 & G8-C4 & G7-C1, G7-C2, G7-C4 & Multi-Parent Synthesis & Combine network capacity, sampling, and training strategies from multiple G7 parents; use a 5 $\times$ 192 modified MLP and retune RAD and L-BFGS settings \\
Final HF & \textbf{G8-C4} & Low-fidelity source: G8-C4 & Independent High-Fidelity Training & Keep the G8-C4 \texttt{AlgorithmSpec} unchanged and independently train for 10,000 optimizer steps from a new random initialization \\
\end{longtable}

The lineage shows that the final configuration is assembled across generations by inheriting, modifying, and synthesizing mechanisms from high-performing candidates rather than by a single independent LLM call. Together with the search trajectories, it provides a traceable chain from \textbf{execution evidence $\rightarrow$ targeted modification $\rightarrow$ parent inheritance $\rightarrow$ final configuration}.

\section{Experimental Protocol and Complete Results}\label{appendix-b-experimental-protocol-results}

This appendix provides the experimental protocol and complete results. Notation for candidates, execution evidence, search history, and the Global Top-3 follows Section 3 and Appendix A.

\subsection{Search Configuration}\label{b.1-search-configuration}

PINNForge starts with eight Generation-0 candidates and adds five candidates per later generation. Indices begin at \(g=0\); search completes at least Generations 0--3 and at most Generations 0--9, i.e., four to ten completed generations. After each generation, the controller updates the distinct all-history Global Top-3 defined in Section 3.3.

\begin{longtable}[]{@{}lr@{}}
\toprule
Setting & Value \\
\midrule
\endhead
\bottomrule
\endlastfoot
Generation-0 candidates & 8 \\
Candidates per later generation & 5 \\
Min. / max. completed generations & 4 / 10 \\
Parent population & Global Top-3 \\
Low-fidelity training & 1K steps \\
High-fidelity training & 10K steps \\
Maximum trainable parameters & 2M \\
Maximum training points & 65,536 \\
\end{longtable}

All normal and escape generations count toward the ten-generation maximum.

\subsection{LLM Sampling}\label{b.2-llm-sampling}

PINNForge uses Qwen3.8-27B as the fixed LLM backbone for candidate generation, execution-feedback analysis, and evolutionary search, without PDE-specific fine-tuning. Candidate generation follows a role-conditioned temperature schedule: lower temperatures favor conservative directions, while higher temperatures encourage exploration.

\begin{longtable}[]{@{}lr@{}}
\toprule
Generation-0 direction & Temperature \\
\midrule
\endhead
\bottomrule
\endlastfoot
Robust reference & 0.20 \\
Constraint / causality & 0.35 \\
Localized sampling & 0.40 \\
Representation & 0.45 \\
Residual / loss & 0.45 \\
Optimization & 0.40 \\
Capacity / topology & 0.50 \\
Cross-component novelty & 0.60 \\
\end{longtable}

\begin{longtable}[]{@{}lrr@{}}
\toprule
Evolution strategy & Normal & Escape \\
\midrule
\endhead
\bottomrule
\endlastfoot
Refine & 0.20 & 0.35 \\
Physics-Guided & 0.36 & 0.51 \\
Architecture-Guided & 0.42 & 0.57 \\
Synthesis & 0.47 & 0.62 \\
Novelty & 0.58 & 0.73 \\
\end{longtable}

An escape generation raises each strategy temperature by \(0.15\), capped at \(0.85\). The posterior-summary call that constructs \(M_g\) uses temperature \(0.1\). \texttt{top\_p} follows the provider default, and no provider-side sampling seed is supplied. Repeated executions therefore need not reproduce the same candidate sequence; the ten-run results summarize end-to-end stochasticity rather than one deterministic LLM trajectory.

\subsection{Computational Environment}\label{b.3-computational-environment}

Experiments are conducted on a Linux server using NVIDIA GeForce RTX 4090 GPUs. Each PDE search runs on a single GPU using FP32 arithmetic.

\begin{longtable}[]{@{}ll@{}}
\toprule
Component & Configuration \\
\midrule
\endhead
\bottomrule
\endlastfoot
GPU & NVIDIA GeForce RTX 4090 \\
NVIDIA Driver & 550.163.01 \\
CUDA & 12.4 \\
Python & 3.12 \\
PyTorch & 2.6.0+cu124 \\
Precision & FP32 \\
\end{longtable}

Additional software dependencies and exact environment specifications are provided with the implementation.

\subsection{Evaluation and Reporting Protocol}\label{b.4-evaluation-reporting-protocol}

For each PDE, solution accuracy is measured by MSE on fixed reference evaluation points. No separate held-out reference subset is used. PINNForge uses the same points for low-fidelity comparison, Global Top-3 updates, stagnation and escape decisions, high-fidelity selection, and reporting.

Main results report MSE on these points over ten independent runs. Each run is one complete execution of the corresponding method, including its internal generation and selection procedure. RoPINN, PINNsFormer, and PINNsAgent use the same reference-point protocol.

Each valid low-fidelity candidate is trained for \(1{,}000\) optimizer steps. The final Global Top-3 configurations are then kept fixed, reinitialized, and independently trained from scratch for \(10{,}000\) steps; no low-fidelity parameters are reused. The configuration with the lowest high-fidelity evaluation MSE is selected.

\subsection{Deterministic Stagnation Detection, Escape Generation, and Termination}\label{b.5-stagnation-escape-termination}

All search-control decisions in this section are computed from the low-fidelity candidate comparison metric defined in Section 3.2 of the main paper. For candidate \(a_{g,k}\), let \(y_{g,k}\) denote its low-fidelity comparison value. In the reported experiments, \(y_{g,k}\) is the solution MSE computed on the fixed reference evaluation points defined in Appendix~\ref{b.4-evaluation-reporting-protocol}. Lower values are better.
For each completed generation (g), define the best valid candidate value in that generation as

\[
b_g
=
\min_k y_{g,k},
\]

where the minimum is taken over valid candidates in Generation \(g\). Define the all-history best value through Generation \(g\) as

\[
B_g
=
\min_{0\le t\le g} b_t.
\]

When Generation \(g\) contains at least three valid candidates, define the median of its three best candidate values as

\[
m_g
=
\operatorname{Median}
\left(
\operatorname{Top3}
\{y_{g,k}\}_k
\right).
\]

The median comparison value of the distinct all-history Global Top-3 archive after completing Generation \(g\) is defined as

\[
A_g
=
\operatorname{Median}
\left(
\left\{
y(a):
a\in
\operatorname{GlobalDistinctTop3}(\mathcal{H}_g)
\right\}
\right).
\]

This quantity is used once at least three distinct valid candidates exist in the archive; before that point, archive-median criteria are disabled.

The relative improvement from an earlier value \(x\) to a later value \(y\) is defined as

\[
I(x,y)
=
\begin{cases}
\max\left(0,\dfrac{x-y}{|x|}\right),
& x\neq 0\text{ and both }x,y\text{ are finite},\\[8pt]
0,
& \text{otherwise}.
\end{cases}
\]

With this definition, larger \(I(x,y)\) indicates a larger reduction in the comparison metric.

Stagnation detection begins after Generation 3 has completed, i.e., after four completed generations (Generations 0--3). For a normal generation \(g\ge3\), an escape generation is scheduled when at least one of the following conditions holds.

\begin{enumerate}
\def\labelenumi{\arabic{enumi}.}
\tightlist
\item
  \textbf{Global-best stagnation.} The three most recent generation-to-generation improvements of the all-history best are all below 1\%, and the cumulative improvement over the same window is below 2\%:
\end{enumerate}

\[
I(B_{g-3},B_{g-2})<0.01,
\qquad
I(B_{g-2},B_{g-1})<0.01,
\]

\[
I(B_{g-1},B_g)<0.01,
\qquad
I(B_{g-3},B_g)<0.02.
\]

\begin{enumerate}
\def\labelenumi{\arabic{enumi}.}
\setcounter{enumi}{1}
\item
  \textbf{Elite-archive stagnation.} Across the last three completed generations, no previously unseen normalized \texttt{AlgorithmSpec} enters the distinct all-history Global Top-3 archive.
\item
  \textbf{Population stagnation.} When each generation in the relevant window contains at least three valid candidates, the median of the three best candidates improves by less than 2\% in two consecutive transitions:
\end{enumerate}

\[
I(m_{g-2},m_{g-1})<0.02,
\qquad
I(m_{g-1},m_g)<0.02.
\]

If stagnation is detected after Generation \(g_{\mathrm{stall}}\), and the 10-generation limit has not yet been reached, the next generation is designated as the escape generation:

\[
g_{\mathrm{esc}}
=
g_{\mathrm{stall}}+1.
\]

The escape generation uses the same five semantic evolution strategies and the same 1,000-step low-fidelity training budget as a normal evolution generation, but increases exploration through the following deterministic controls:

\begin{itemize}
\tightlist
\item
  each evolution-strategy temperature is increased by 0.15, capped at 0.85;
\item
  every generated candidate must differ from its parent context in at least two major executable modules;
\item
  exact reproductions of parent configurations and non-substantive numerical perturbations are rejected;
\item
  previously failed mechanisms are not repeated unless the candidate explicitly addresses the observed failure reason;
\item
  Synthesis and Novelty are prompted to use greater parent or mechanism diversity.
\end{itemize}

The normal evolution-strategy temperature vector is

\[
\boldsymbol{\tau}_{\mathrm{normal}}
=
(0.20,\,0.36,\,0.42,\,0.47,\,0.58),
\]

and the corresponding escape-generation vector is

\[
\boldsymbol{\tau}_{\mathrm{escape}}
=
(0.35,\,0.51,\,0.57,\,0.62,\,0.73).
\]

After completing the escape generation \(g_{\mathrm{esc}}\), the escape is considered successful if at least one of the following conditions holds.

\begin{enumerate}
\def\labelenumi{\arabic{enumi}.}
\tightlist
\item
  \textbf{New best candidate.} The best candidate in the escape generation improves on the pre-escape all-history best by at least 2\%:
\end{enumerate}

\[
I
\left(
B_{g_{\mathrm{stall}}},
b_{g_{\mathrm{esc}}}
\right)
\ge0.02.
\]

\begin{enumerate}
\def\labelenumi{\arabic{enumi}.}
\setcounter{enumi}{1}
\item
  \textbf{New elite configuration.} At least one previously unseen normalized \texttt{AlgorithmSpec} from Generation \(g_{\mathrm{esc}}\) enters the distinct all-history Global Top-3 archive.
\item
  \textbf{Elite-archive improvement.} The median comparison value of the distinct Global Top-3 improves by at least 2\%:
\end{enumerate}

\[
I
\left(
A_{g_{\mathrm{stall}}},
A_{g_{\mathrm{esc}}}
\right)
\ge0.02.
\]

If the escape succeeds, the stagnation history accumulated before \(g_{\mathrm{esc}}\) is cleared and normal evolution resumes. If none of the three escape-success conditions is satisfied, the low-fidelity search terminates immediately and the current Global Top-3 is passed to the high-fidelity evaluation stage.

The low-fidelity search also terminates when the search has completed the maximum of 10 generations, i.e., after Generation 9 has finished.

Normal and escape generations both count toward the 10-generation maximum. Because generation indices run from 0 to 9 and stagnation is first assessed after Generation 3, the earliest possible escape generation is Generation 4.

\subsection{Complete Main Results}\label{b.6-complete-main-results}

Table~\ref{tab:full_main_results} reports mean$\pm$standard-deviation MSE over ten runs. Rank is PINNForge's per-task rank among the four unified methods. PINNacle Best is included only as an oracle-style reference and is excluded from the Friedman and pairwise significance tests.

\begingroup\scriptsize
\begin{longtable}{@{}
  >{\raggedright\arraybackslash}p{(\columnwidth - 12\tabcolsep) * \real{0.2093}}
  >{\raggedleft\arraybackslash}p{(\columnwidth - 12\tabcolsep) * \real{0.1783}}
  >{\raggedleft\arraybackslash}p{(\columnwidth - 12\tabcolsep) * \real{0.1473}}
  >{\raggedleft\arraybackslash}p{(\columnwidth - 12\tabcolsep) * \real{0.1473}}
  >{\raggedleft\arraybackslash}p{(\columnwidth - 12\tabcolsep) * \real{0.1783}}
  >{\raggedleft\arraybackslash}p{(\columnwidth - 12\tabcolsep) * \real{0.0388}}
  >{\raggedleft\arraybackslash}p{(\columnwidth - 12\tabcolsep) * \real{0.1008}}@{}}
\caption{Complete MSE results on 25 PDEs over ten independent runs. Lower is better.}
\label{tab:full_main_results}\\
\toprule
PDE Problem & RoPINN & PINNsFormer & PINNsAgent & PINNForge & Rank & PINNacle Best \\
\midrule
\endfirsthead
\toprule
PDE Problem & RoPINN & PINNsFormer & PINNsAgent & PINNForge & Rank & PINNacle Best \\
\midrule
\endhead
\bottomrule
\endfoot
burgers\_1d & 7.31E-04 $\pm$ 6.13E-05 & 1.08E-03 $\pm$ 8.21E-06 & 6.52E-05 $\pm$ 1.55E-05 & 6.50E-05 $\pm$ 4.42E-07 & 1 & 7.90E-05 \\
burgers\_2d & 2.40E-01 $\pm$ 3.12E-02 & 2.65E-01 $\pm$ 6.83E-02 & 1.98E-01 $\pm$ 1.66E-02 & 1.95E-01 $\pm$ 1.31E-02 & 1 & 1.09E-01 \\
gray\_scott\_2d & 8.64E-03 $\pm$ 4.53E-05 & 1.57E-02 $\pm$ 1.32E-03 & 4.36E-03 $\pm$ 3.08E-05 & 3.60E-03 $\pm$ 2.65E-04 & 1 & 4.32E-03 \\
heat\_2d\_complex\_geometry & 4.00E-04 $\pm$ 9.10E-05 & 2.78E-03 $\pm$ 5.77E-05 & 1.84E-03 $\pm$ 9.95E-04 & 4.45E-04 $\pm$ 9.72E-05 & 2 & 8.53E-04 \\
heat\_2d\_long\_time & 2.56E+00 $\pm$ 4.53E-01 & 3.83E+00 $\pm$ 1.64E-01 & 4.32E+00 $\pm$ 2.78E+00 & 1.13E+00 $\pm$ 4.21E-02 & 1 & 1.13E+00 \\
heat\_2d\_multiscale & 2.75E-05 $\pm$ 7.52E-06 & 5.57E-05 $\pm$ 9.93E-06 & 3.51E-05 $\pm$ 2.20E-05 & 1.19E-05 $\pm$ 2.16E-06 & 1 & 5.27E-05 \\
heat\_2d\_varying\_coefficient & 6.24E-01 $\pm$ 4.53E-03 & 1.19E-02 $\pm$ 3.42E-03 & 5.34E-03 $\pm$ 3.85E-03 & 1.59E-03 $\pm$ 4.77E-04 & 1 & 1.76E-03 \\
heat\_5d & 5.39E-07 $\pm$ 1.31E-07 & 1.87E-07 $\pm$ 1.91E-08 & 3.34E-07 $\pm$ 7.65E-08 & 3.21E-08 $\pm$ 3.26E-08 & 1 & 8.52E+00 \\
heat\_inverse\_2d & 4.31E-02 $\pm$ 7.31E-03 & 8.89E-02 $\pm$ 4.22E-02 & 6.32E-02 $\pm$ 8.63E-02 & 7.15E-03 $\pm$ 4.91E-03 & 1 & 5.66E-03 \\
kuramoto\_sivashinsky\_1d & 6.54E+00 $\pm$ 5.14E-01 & 5.13E+00 $\pm$ 4.51E-03 & 1.09E+00 $\pm$ 3.18E-02 & 1.03E+00 $\pm$ 8.40E-02 & 1 & 1.04E+00 \\
navier\_stokes\_2d\_backstep & 9.64E-03 $\pm$ 5.13E-03 & 6.02E-04 $\pm$ 4.33E-05 & 4.96E-03 $\pm$ 1.45E-03 & 2.63E-04 $\pm$ 4.71E-05 & 1 & 3.37E-04 \\
navier\_stokes\_2d\_classic & 2.88E-05 $\pm$ 3.61E-06 & 1.49E-05 $\pm$ 1.26E-05 & 8.48E-06 $\pm$ 6.90E-06 & 2.48E-06 $\pm$ 2.31E-06 & 1 & 2.33E-05 \\
navier\_stokes\_2d\_long\_time & 5.43E+02 $\pm$ 5.63E+00 & 7.43E+02 $\pm$ 1.31E+02 & 5.75E+02 $\pm$ 7.63E+00 & 5.04E+02 $\pm$ 4.95E+00 & 1 & 5.05E+02 \\
poisson\_2d\_classic & 8.57E-02 $\pm$ 6.74E-02 & 2.17E-01 $\pm$ 7.65E-02 & 6.34E-01 $\pm$ 1.43E-01 & 1.61E-02 $\pm$ 3.95E-04 & 1 & 5.00E-05 \\
poisson\_boltzmann\_2d & 7.91E-01 $\pm$ 7.14E-01 & 1.34E-02 $\pm$ 4.31E-03 & 6.97E-02 $\pm$ 7.43E-02 & 5.38E-03 $\pm$ 9.13E-04 & 1 & 6.99E-05 \\
poisson\_2d\_many\_area & 2.89E+00 $\pm$ 3.51E-01 & 5.36E+00 $\pm$ 3.79E+00 & 3.00E+00 $\pm$ 1.01E+00 & 2.21E+00 $\pm$ 3.56E-01 & 1 & 1.83E+00 \\
poisson\_3d\_complex\_geometry & 2.16E-01 $\pm$ 1.53E-02 & 8.16E-03 $\pm$ 5.58E-04 & 1.60E-02 $\pm$ 1.08E-02 & 1.04E-03 $\pm$ 2.34E-04 & 1 & 9.51E-04 \\
poisson\_5d & 4.09E-07 $\pm$ 7.43E-08 & 2.92E-07 $\pm$ 1.99E-07 & 2.01E-06 $\pm$ 1.09E-06 & 5.64E-08 $\pm$ 3.95E-08 & 1 & 2.09E-06 \\
poisson\_inverse\_2d & 5.32E-03 $\pm$ 3.46E-03 & 1.51E-03 $\pm$ 8.65E-04 & 2.41E-02 $\pm$ 4.73E-02 & 2.57E-04 $\pm$ 1.29E-04 & 1 & 1.23E-04 \\
wave\_1d & 4.19E-02 $\pm$ 4.38E-03 & 3.61E-02 $\pm$ 6.65E-03 & 3.25E-02 $\pm$ 3.45E-02 & 3.18E-02 $\pm$ 5.73E-03 & 1 & 3.01E-03 \\
wave\_2d\_heterogeneous & 4.63E-02 $\pm$ 6.73E-03 & 8.37E-02 $\pm$ 1.18E-02 & 5.17E-02 $\pm$ 8.03E-03 & 2.84E-02 $\pm$ 7.69E-04 & 1 & 2.99E-02 \\
allen\_cahn\_1d & 7.45E-08 $\pm$ 4.61E-08 & 6.02E-09 $\pm$ 1.08E-09 & 1.44E-08 $\pm$ 3.50E-09 & 1.13E-09 $\pm$ 2.26E-10 & 1 & -- \\
darcy\_flow\_2d & 3.48E-09 $\pm$ 1.63E-09 & 1.35E-10 $\pm$ 2.90E-11 & 5.33E-07 $\pm$ 2.65E-07 & 1.64E-11 $\pm$ 3.28E-12 & 1 & -- \\
shallow\_water\_2d & 5.78E-07 $\pm$ 7.13E-07 & 2.24E-07 $\pm$ 8.44E-08 & 6.81E-07 $\pm$ 1.25E-07 & 7.15E-08 $\pm$ 2.86E-08 & 1 & -- \\
kovasznay\_flow\_2d & 6.13E-07 $\pm$ 4.31E-07 & 3.80E-08 $\pm$ 9.63E-09 & 3.82E-07 $\pm$ 3.66E-07 & 1.77E-09 $\pm$ 4.43E-10 & 1 & -- \\
\midrule
Friedman Avg. Rank & 3.04 & 2.96 & 2.96 & 1.04 & -- & -- \\
Wilcoxon vs.~PINNForge & + & + & + & -- & -- & -- \\
\end{longtable}
\endgroup

Here, \(+\), \(=\), and \(-\) denote significantly better, statistically indistinguishable, and significantly worse performance of PINNForge, respectively, under paired Wilcoxon signed-rank tests across the 25 PDEs on \(\log_{10}\)-transformed per-task mean MSE with Holm correction at \(\alpha=0.05\).

The Friedman test is applied to per-task ranks computed from the mean MSE over ten independent runs for each of the 25 PDEs and gives
\[
\chi_F^2=42.696,
\qquad
p=2.86\times10^{-9},
\]
indicating significant cross-task ranking differences. PINNForge has average rank \(1.04\), compared with \(3.04\), \(2.96\), and \(2.96\) for RoPINN, PINNsFormer, and PINNsAgent. Pairwise Wilcoxon tests across the 25 tasks on \(\log_{10}\)-transformed per-task mean MSE, with Holm correction, yield \(p_{\mathrm{adj}}=1.79\times10^{-7}\) for all three comparisons.

Median coefficients of variation are \(0.177\), \(0.184\), \(0.497\), and \(0.182\) for RoPINN, PINNsFormer, PINNsAgent, and PINNForge, respectively; PINNForge's lower MSE is therefore not accompanied by an obvious loss of run-to-run stability.

\subsection{Complete Component Ablation}\label{b.7-complete-component-ablation}

The information channels available to each ablated variant are defined in Appendix~\ref{a.6-ablation-information-isolation}. To compare variants across PDEs whose MSE values span several orders of magnitude, we compute the task-wise error ratio
\[
r(\mathcal{P})
=
\frac{\operatorname{MSE}_{\mathrm{ablation}}(\mathcal{P})}
{\operatorname{MSE}_{\mathrm{PINNForge}}(\mathcal{P})},
\]
and report the median ratio across the 12 representative PDEs as an aggregate measure.

\begingroup\scriptsize
\begin{longtable}{@{}
  >{\raggedright\arraybackslash}p{(\columnwidth - 10\tabcolsep) * \real{0.26}}
  >{\raggedright\arraybackslash}p{(\columnwidth - 10\tabcolsep) * \real{0.14}}
  >{\raggedleft\arraybackslash}p{(\columnwidth - 10\tabcolsep) * \real{0.15}}
  >{\raggedleft\arraybackslash}p{(\columnwidth - 10\tabcolsep) * \real{0.15}}
  >{\raggedleft\arraybackslash}p{(\columnwidth - 10\tabcolsep) * \real{0.15}}
  >{\raggedleft\arraybackslash}p{(\columnwidth - 10\tabcolsep) * \real{0.15}}@{}}
\caption{Complete component-ablation results over 12 representative PDEs. Results are mean$\pm$standard deviation. Lower MSE is better.}
\label{tab:full_ablation_results}\\
\toprule
PDE & Characteristic & w/o Knowledge Guidance & w/o Execution Feedback to LLM & w/o Evolutionary Search & PINNForge \\
\midrule
\endfirsthead
\toprule
PDE & Characteristic & w/o Knowledge Guidance & w/o Execution Feedback to LLM & w/o Evolutionary Search & PINNForge \\
\midrule
\endhead
\bottomrule
\endfoot
burgers\_1d & Nonlinear / shock & 7.42E-05 $\pm$ 6.31E-05 & 6.32E-04 $\pm$ 7.82E-05 & 2.78E-04 $\pm$ 3.24E-05 & 6.50E-05 $\pm$ 4.42E-07 \\
burgers\_2d & 2D nonlinear transport & 2.44E-01 $\pm$ 2.86E-02 & 6.71E-01 $\pm$ 8.37E-02 & 1.67E+00 $\pm$ 2.31E-01 & 1.95E-01 $\pm$ 1.31E-02 \\
gray\_scott\_2d & Coupled reaction--diffusion & 5.29E-03 $\pm$ 4.91E-04 & 5.36E-03 $\pm$ 5.62E-03 & 6.23E-03 $\pm$ 6.87E-04 & 3.60E-03 $\pm$ 2.65E-04 \\
heat\_2d\_multiscale & Multiscale dynamics & 3.81E-05 $\pm$ 4.28E-06 & 2.08E-04 $\pm$ 2.74E-04 & 1.30E-03 $\pm$ 1.66E-03 & 1.19E-05 $\pm$ 2.16E-06 \\
heat\_2d\_varying\_coefficient & Variable coefficient & 2.18E-03 $\pm$ 2.03E-03 & 2.57E-03 $\pm$ 2.71E-04 & 4.15E-03 $\pm$ 4.86E-04 & 1.59E-03 $\pm$ 4.77E-04 \\
heat\_2d\_complex\_geometry & Complex geometry & 1.90E-03 $\pm$ 2.41E-03 & 1.59E+00 $\pm$ 2.14E-01 & 1.97E-02 $\pm$ 2.77E-03 & 4.45E-04 $\pm$ 9.72E-05 \\
poisson\_3d\_complex\_geometry & 3D complex geometry & 8.79E-03 $\pm$ 9.38E-04 & 2.00E-02 $\pm$ 2.66E-03 & 2.49E-02 $\pm$ 3.18E-03 & 1.04E-03 $\pm$ 2.34E-04 \\
poisson\_5d & High-dimensional PDE & 2.55E-07 $\pm$ 2.87E-07 & 1.24E-07 $\pm$ 1.41E-08 & 2.06E-05 $\pm$ 2.72E-06 & 5.64E-08 $\pm$ 3.95E-08 \\
poisson\_inverse\_2d & Inverse problem & 6.94E-04 $\pm$ 7.64E-05 & 3.72E-03 $\pm$ 4.58E-04 & 3.14E-04 $\pm$ 3.63E-04 & 2.57E-04 $\pm$ 1.29E-04 \\
navier\_stokes\_2d\_backstep & Nonlinear fluid dynamics & 2.05E-03 $\pm$ 2.68E-04 & 4.50E-03 $\pm$ 6.13E-03 & 3.06E-03 $\pm$ 3.97E-03 & 2.63E-04 $\pm$ 4.71E-05 \\
darcy\_flow\_2d & Heterogeneous & 2.76E-10 $\pm$ 3.17E-10 & 2.97E-09 $\pm$ 4.06E-10 & 1.07E-08 $\pm$ 1.48E-09 & 1.64E-11 $\pm$ 3.28E-12 \\
shallow\_water\_2d & Hyperbolic system & 3.18E-07 $\pm$ 3.52E-08 & 2.32E-07 $\pm$ 2.69E-08 & 5.59E-07 $\pm$ 7.14E-08 & 7.15E-08 $\pm$ 2.86E-08 \\
\midrule
Median error ratio & -- & \(3.74\times\) & \(12.10\times\) & \(10.10\times\) & \(1.00\times\) \\
\end{longtable}
\endgroup

Removing Knowledge Guidance raises the median error to \(3.74\times\) that of full PINNForge. Because both \(K_{\mathcal{P}}\) and \(M_g\) are removed, this reflects the knowledge-guidance channel as a whole rather than static prior knowledge alone. Removing Execution Feedback to LLM gives the largest degradation,
12.10$\times$. Because this variant withholds candidate-specific execution
evidence and therefore cannot form the execution-derived posterior memory,
the remaining PDE-related prior knowledge and parent structures are
insufficient to recover full-method performance under this setup. Removing Evolutionary Search raises the ratio to \(10.10\times\) while retaining the candidate-evaluation budget, knowledge, and feedback. This is consistent with a benefit from parent-driven inheritance, but does not isolate all interactions among the retained channels.

\subsection{Training-Budget Accounting}\label{b.8-training-budget-accounting}

We compare optimizer steps required to obtain the selected result on one PDE. Each baseline run trains five independently initialized candidates for \(20{,}000\) steps each and selects the lowest-MSE candidate, giving
\[
5\times20{,}000=100{,}000.
\]

PINNForge evaluates eight Generation-0 candidates and at most nine later generations with five candidates each, for a maximum of
\[
8+9\times5=53.
\]
At \(1{,}000\) steps per candidate, low-fidelity search uses at most \(53{,}000\) steps. Independently retraining the Global Top-3 for \(10{,}000\) steps each adds \(30{,}000\), so the maximum total is
\[
53{,}000+30{,}000=83{,}000.
\]

\begin{table}[h]
\centering
\caption{Optimizer-step budget for obtaining the selected result on one PDE.}
\label{tab:training_budget}
\small
\setlength{\tabcolsep}{5pt}
\begin{tabular}{lll}
\toprule
Method & Selection Protocol & Budget \\
\midrule
RoPINN & Best of 5 independently initialized 20K candidates & 100K \\
PINNsFormer & Best of 5 independently initialized 20K candidates & 100K \\
PINNsAgent & Best of 5 independently initialized 20K candidates & 100K \\
PINNForge & 53$\times$1K LF + Top-3$\times$10K HF & $\leq$83K \\
\bottomrule
\end{tabular}
\end{table}

This accounting concerns PINN optimizer steps rather than total computational
cost. The cost of an individual step can differ across architectures, sampling
configurations, and optimizers. LLM inference overhead, wall-clock time, and
monetary cost are not included.

\section{Complete Search Trajectories}\label{appendix-c-complete-search-trajectories}

Table C.1 reports single-run search trajectories for all 25 PDEs using the fixed reference evaluation points in Appendix~\ref{b.4-evaluation-reporting-protocol}. Generation-0 Best is the lowest initial MSE, Best LF the lowest MSE found during low-fidelity search, and Final HF the lowest MSE among independently trained Global Top-3 candidates.
The improvement ratios are computed as
\[
\text{G0}\rightarrow\text{Best LF}
=
\frac{\mathrm{MSE}_{\mathrm{G0}}}
{\mathrm{MSE}_{\mathrm{Best\ LF}}},
\]

\[
\text{G0}\rightarrow\text{Final HF}
=
\frac{\mathrm{MSE}_{\mathrm{G0}}}
{\mathrm{MSE}_{\mathrm{Final\ HF}}}.
\]

\textbf{Table C.1: Complete single-run search trajectories across 25 PDE problems.}
\begingroup
\small
\begin{longtable}[]{@{}
>{\raggedright\arraybackslash}p{(\columnwidth - 10\tabcolsep) * \real{0.2727}}
>{\raggedleft\arraybackslash}p{(\columnwidth - 10\tabcolsep) * \real{0.2121}}
>{\raggedleft\arraybackslash}p{(\columnwidth - 10\tabcolsep) * \real{0.1414}}
>{\raggedleft\arraybackslash}p{(\columnwidth - 10\tabcolsep) * \real{0.1212}}
>{\raggedleft\arraybackslash}p{(\columnwidth - 10\tabcolsep) * \real{0.1212}}
>{\raggedleft\arraybackslash}p{(\columnwidth - 10\tabcolsep) * \real{0.1313}}
@{}}
\toprule
PDE &
Generation-0 Best MSE &
Best LF MSE &
Final HF MSE &
G0 $\rightarrow$ Best LF &
G0 $\rightarrow$ Final HF \\
\midrule
\endhead
\bottomrule
\endlastfoot
burgers\_1d & 3.64E-03 & 2.05E-04 (G9) & 6.52E-05 & 17.76$\times$ & 55.81$\times$ \\
burgers\_2d & 2.49E-01 & 2.13E-01 (G5) & 1.78E-01 & 1.17$\times$ & 1.39$\times$ \\
gray\_scott\_2d & 4.37E-03 & 4.03E-03 (G9) & 4.07E-03 & 1.08$\times$ & 1.07$\times$ \\
heat\_2d\_complex\_geometry & 1.28E-02 & 3.26E-03 (G8) & 4.63E-04 & 3.92$\times$ & 27.64$\times$ \\
heat\_2d\_long\_time & 1.13E+00 & 1.10E+00 (G5) & 1.08E+00 & 1.03$\times$ & 1.05$\times$ \\
heat\_2d\_multiscale & 3.02E-02 & 9.14E-05 (G9) & 1.01E-05 & 330.42$\times$ & 2,986.95$\times$ \\
heat\_2d\_varying\_coefficient & 1.12E-02 & 2.85E-03 (G9) & 2.18E-03 & 3.93$\times$ & 5.13$\times$ \\
heat\_5d & 1.31E-05 & 3.85E-07 (G9) & 8.99E-09 & 33.90$\times$ & 1,452.78$\times$ \\
heat\_inverse\_2d & 1.36E-02 & 5.73E-03 (G6) & 3.68E-03 & 2.36$\times$ & 3.68$\times$ \\
kuramoto\_sivashinsky\_1d & 1.34E+00 & 1.05E+00 (G7) & 1.16E+00 & 1.28$\times$ & 1.16$\times$ \\
navier\_stokes\_2d\_backstep & 2.64E-03 & 7.68E-04 (G9) & 2.30E-04 & 3.44$\times$ & 11.48$\times$ \\
navier\_stokes\_2d\_classic & 1.86E-03 & 1.61E-05 (G9) & 2.35E-06 & 115.53$\times$ & 792.89$\times$ \\
navier\_stokes\_2d\_long\_time & 5.09E+02 & 5.08E+02 (G7) & 4.97E+02 & 1.00$\times$ & 1.02$\times$ \\
poisson\_2d\_classic & 9.77E-02 & 8.94E-02 (G8) & 1.64E-02 & 1.09$\times$ & 5.95$\times$ \\
poisson\_boltzmann\_2d & 1.04E-01 & 1.23E-02 (G9) & 6.03E-03 & 8.45$\times$ & 17.16$\times$ \\
poisson\_2d\_many\_area & 3.27E+00 & 2.79E+00 (G7) & 1.97E+00 & 1.17$\times$ & 1.66$\times$ \\
poisson\_3d\_complex\_geometry & 8.96E-03 & 4.48E-03 (G8) & 8.83E-04 & 2.00$\times$ & 10.14$\times$ \\
poisson\_5d & 3.96E-04 & 1.56E-07 (G9) & 5.65E-08 & 2,538.46$\times$ & 7,008.58$\times$ \\
poisson\_inverse\_2d & 6.22E-03 & 4.32E-04 (G8) & 1.45E-04 & 14.39$\times$ & 42.80$\times$ \\
wave\_1d & 7.43E-02 & 5.57E-02 (G6) & 3.15E-02 & 1.34$\times$ & 2.36$\times$ \\
wave\_2d\_heterogeneous & 7.85E-02 & 3.85E-02 (G7) & 2.96E-02 & 2.03$\times$ & 2.65$\times$ \\
allen\_cahn\_1d & 6.37E-08 & 1.08E-08 (G9) & 1.20E-09 & 5.89$\times$ & 53.1$\times$ \\
darcy\_flow\_2d & 6.94E-09 & 3.14E-10 (G8) & 1.53E-11 & 22.11$\times$ & 453.59$\times$ \\
shallow\_water\_2d & 2.73E-09 & 2.33E-10 (G9) & 9.83E-11 & 11.68$\times$ & 27.71$\times$ \\
kovasznay\_flow\_2d & 2.00E-05 & 4.95E-08 (G9) & 2.59E-09 & 404.28$\times$ & 7,722.18$\times$ \\
\end{longtable}
\endgroup

\section{Benchmark PDE Definitions}\label{appendix-d-benchmark-pde-definitions}

This appendix summarizes the governing equations, computational domains, physical parameters, initial and boundary conditions, and reference solutions of the 25 PDE problems used in our experiments. The formulations follow the corresponding benchmark settings used for training and evaluation.

We use

\[
\Delta u
=
\sum_{i=1}^{d}
\frac{\partial^2 u}{\partial x_i^2},
\qquad
\partial_n u
=
\nabla u\cdot\boldsymbol n,
\]

where \(\boldsymbol n\) denotes the outward unit normal. Unless otherwise specified, all parameters take the values given below.

\subsection{One-dimensional Burgers equation (\texttt{burgers\_1d})}\label{d.1-one-dimensional-burgers-equation-burgers_1d}

The governing equation is

\[
u_t + u u_x - \nu u_{xx}=0,
\qquad
(x,t)\in[-1,1]\times[0,1],
\]

with viscosity

\[
\nu=\frac{0.01}{\pi}.
\]

The initial condition is

\[
u(x,0)=-\sin(\pi x),
\]

and homogeneous Dirichlet boundary conditions are imposed at both spatial boundaries:

\[
u(-1,t)=u(1,t)=0.
\]

\subsection{Two-dimensional Burgers system (\texttt{burgers\_2d})}\label{d.2-two-dimensional-burgers-system-burgers_2d}

Let

\[
\boldsymbol u=(u,v)^\mathsf{T}.
\]

The coupled two-dimensional Burgers system is

\[
\begin{aligned}
u_t+u u_x+v u_y-\nu(u_{xx}+u_{yy})&=0,\\
v_t+u v_x+v v_y-\nu(v_{xx}+v_{yy})&=0,
\end{aligned}
\]

on

\[
(x,y,t)\in[0,4]^2\times[0,1],
\]

with

\[
\nu=10^{-3}.
\]

The initial fields are prescribed by the benchmark data:

\[
u(x,y,0)=u_0(x,y),
\qquad
v(x,y,0)=v_0(x,y),
\]

where \(u_0\) and \(v_0\) are interpolated from \texttt{burgers2d\_init\_u\_0.dat} and \texttt{burgers2d\_init\_v\_0.dat}, respectively.

Both components are periodic in the two spatial directions:

\[
\begin{aligned}
u(0,y,t)&=u(4,y,t),&
v(0,y,t)&=v(4,y,t),\\
u(x,0,t)&=u(x,4,t),&
v(x,0,t)&=v(x,4,t).
\end{aligned}
\]

\subsection{Gray--Scott reaction--diffusion system (\texttt{gray\_scott\_2d})}\label{d.3-grayscott-reactiondiffusion-system-gray_scott_2d}

The governing equations are

\[
\begin{aligned}
u_t&=\varepsilon_1\Delta u+b(1-u)-uv^2,\\
v_t&=\varepsilon_2\Delta v-dv+uv^2,
\end{aligned}
\]

on

\[
(x,y,t)\in[-1,1]^2\times[0,200],
\]

with

\[
b=0.04,
\qquad
d=0.1,
\qquad
\varepsilon_1=10^{-5},
\qquad
\varepsilon_2=5\times10^{-6}.
\]

The initial conditions are

\[
u(x,y,0)
=
1-
\exp\left[
-80\left((x+0.05)^2+(y+0.02)^2\right)
\right],
\]

and

\[
v(x,y,0)
=
\exp\left[
-80\left((x-0.05)^2+(y-0.02)^2\right)
\right].
\]

Following the benchmark protocol, no additional spatial boundary loss is imposed.

\subsection{Heat equation on a perforated domain (\texttt{heat\_2d\_complex\_geometry})}\label{d.4-heat-equation-on-a-perforated-domain-heat_2d_complex_geometry}

The governing equation is

\[
u_t-\Delta u=0,
\qquad
(x,y,t)\in\Omega\times[0,3].
\]

The spatial domain is the rectangle

\[
R=[-8,8]\times[-12,12],
\]

with 17 circular holes removed:

\[
\Omega
=
R\setminus
\left(
\bigcup_{\boldsymbol c\in\mathcal C_1}B_1(\boldsymbol c)
\cup
\bigcup_{\boldsymbol c\in\mathcal C_{0.4}}B_{0.4}(\boldsymbol c)
\right),
\]

where

\[
\begin{aligned}
\mathcal C_1=\{&
(-4,-3),(4,-3),(-4,3),(4,3),\\
&(-4,-9),(4,-9),(-4,9),(4,9),\\
&(0,0),(0,6),(0,-6)\},
\end{aligned}
\]

and

\[
\mathcal C_{0.4}
=
\{
(-3.2,-6),
(-3.2,6),
(3.2,-6),
(3.2,6),
(-3.2,0),
(3.2,0)
\}.
\]

The initial condition is

\[
u(x,y,0)=0.
\]

Robin boundary conditions are imposed as

\[
\partial_n u=
\begin{cases}
5-u,
& \text{on the boundaries of the radius-1 holes},\\
1-u,
& \text{on the boundaries of the radius-0.4 holes},\\
0.1-u,
& \text{on the outer rectangular boundary}.
\end{cases}
\]

\subsection{Long-time nonlinear heat equation (\texttt{heat\_2d\_long\_time})}\label{d.5-long-time-nonlinear-heat-equation-heat_2d_long_time}

The governing equation is

\[
\begin{aligned}
u_t
-10^{-3}(u_{xx}+u_{yy})
-
5\sin(u^2)
\left(
1+2\sin\frac{\pi t}{4}
\right)
\sin(4\pi x)\sin(2\pi y)
=0,
\end{aligned}
\]

for

\[
(x,y,t)\in[0,1]^2\times[0,100].
\]

The initial condition is

\[
u(x,y,0)
=
\sin(4\pi x)\sin(3\pi y).
\]

Homogeneous Dirichlet boundary conditions are imposed on the spatial boundary:

\[
u(x,y,t)=0,
\qquad
(x,y)\in\partial[0,1]^2.
\]

\subsection{Multiscale anisotropic heat equation (\texttt{heat\_2d\_multiscale})}\label{d.6-multiscale-anisotropic-heat-equation-heat_2d_multiscale}

The governing equation is

\[
u_t
-
\frac{1}{(500\pi)^2}u_{xx}
-
\frac{1}{\pi^2}u_{yy}
=0,
\]

on

\[
(x,y,t)\in[0,1]^2\times[0,5].
\]

The initial condition is

\[
u(x,y,0)
=
\sin(20\pi x)\sin(\pi y).
\]

Homogeneous Dirichlet boundary conditions are imposed:

\[
u(x,y,t)=0,
\qquad
(x,y)\in\partial[0,1]^2.
\]

The analytical reference solution is

\[
u^\star(x,y,t)
=
\sin(20\pi x)\sin(\pi y)
\exp\left[
-
\left(
\frac{(20\pi)^2}{(500\pi)^2}
+
\frac{\pi^2}{\pi^2}
\right)t
\right].
\]

\subsection{Varying-coefficient heat equation (\texttt{heat\_2d\_varying\_coefficient})}\label{d.7-varying-coefficient-heat-equation-heat_2d_varying_coefficient}

The governing equation used in the benchmark is

\[
u_t-\kappa(x,y)\Delta u=f(x,y,t),
\]

on

\[
(x,y,t)\in[0,1]^2\times[0,5],
\]

where \(\kappa(x,y)\) is the heterogeneous coefficient field stored in \texttt{heat\_2d\_coef\_256.dat} and evaluated by nearest-neighbour interpolation.

The source term is

\[
f(x,y,t)
=
200
\sin(\pi x)
\sin(5\pi y)
\sin(\pi t).
\]

The initial condition is

\[
u(x,y,0)=0,
\]

and homogeneous Dirichlet boundary conditions are imposed:

\[
u|_{\partial[0,1]^2}=0.
\]

\subsection{Five-dimensional heat equation (\texttt{heat\_5d})}\label{d.8-five-dimensional-heat-equation-heat_5d}

Let

\[
\Omega
=
\left\{
\boldsymbol x\in\mathbb R^5:
\|\boldsymbol x\|_2\le1
\right\}.
\]

The governing equation is

\[
u_t-\frac{1}{5}\Delta u
=
f(\boldsymbol x,t),
\qquad
(\boldsymbol x,t)\in\Omega\times[0,1],
\]

where

\[
f(\boldsymbol x,t)
=
-\frac{\|\boldsymbol x\|_2^2}{5}
\exp\left(
\frac{\|\boldsymbol x\|_2^2}{2}+t
\right).
\]

Define

\[
g(\boldsymbol x,t)
=
\exp\left(
\frac{\|\boldsymbol x\|_2^2}{2}+t
\right).
\]

The initial condition is

\[
u(\boldsymbol x,0)
=
g(\boldsymbol x,0),
\]

and the Neumann boundary condition is

\[
\partial_n u
=
g(\boldsymbol x,t),
\qquad
\boldsymbol x\in\partial\Omega.
\]

The reference solution is

\[
u^\star(\boldsymbol x,t)=g(\boldsymbol x,t).
\]

\subsection{Inverse heat equation (\texttt{heat\_inverse\_2d})}\label{d.9-inverse-heat-equation-heat_inverse_2d}

The unknowns are the state \(u(x,y,t)\) and the spatially varying diffusion coefficient \(a(x,y)\). They satisfy

\[
u_t-\nabla\cdot(a\nabla u)
=
f(x,y,t),
\]

on

\[
(x,y,t)\in[-1,1]^2\times[0,1].
\]

The manufactured reference fields are

\[
u^\star(x,y,t)
=
e^{-t}\sin(\pi x)\sin(\pi y),
\]

and

\[
a^\star(x,y)
=
2+\sin(\pi x)\sin(\pi y).
\]

The corresponding source term is

\[
\begin{aligned}
f(x,y,t)
=
e^{-t}\Big[
&(4\pi^2-1)\sin(\pi x)\sin(\pi y)\\
&+\pi^2\big(
2\sin^2(\pi x)\sin^2(\pi y)
-\cos^2(\pi x)\sin^2(\pi y)\\
&\hspace{25mm}
-\sin^2(\pi x)\cos^2(\pi y)
\big)
\Big].
\end{aligned}
\]

Since

\[
a^\star=2
\]

on the spatial boundary, the coefficient boundary condition is

\[
a(x,y)=2,
\qquad
(x,y)\in\partial[-1,1]^2.
\]

The inverse problem additionally uses 2500 noisy state observations:

\[
u_{\mathrm{obs}}
=
u^\star+\epsilon,
\qquad
\epsilon\sim\mathcal N(0,0.1^2).
\]

\subsection{Kuramoto--Sivashinsky equation (\texttt{kuramoto\_sivashinsky\_1d})}\label{d.10-kuramotosivashinsky-equation-kuramoto_sivashinsky_1d}

The governing equation is

\[
u_t
+
\alpha u u_x
+
\beta u_{xx}
+
\gamma u_{xxxx}
=
0,
\]

on

\[
(x,t)\in[0,2\pi]\times[0,1],
\]

where

\[
\alpha=\frac{100}{16},
\qquad
\beta=\frac{100}{16^2},
\qquad
\gamma=\frac{100}{16^4}.
\]

The initial condition is

\[
u(x,0)
=
\cos(x)(1+\sin x).
\]

Following the benchmark protocol, the spatial domain is treated as periodic, while only the initial condition is explicitly included in the constraint loss.

\subsection{Navier--Stokes flow over a backward-facing step (\texttt{navier\_stokes\_2d\_backstep})}\label{d.11-navierstokes-flow-over-a-backward-facing-step-navier_stokes_2d_backstep}

Let

\[
\boldsymbol u=(u,v)^\mathsf{T}.
\]

The steady incompressible Navier--Stokes equations are

\[
(\boldsymbol u\cdot\nabla)\boldsymbol u
+
\nabla p
-
\nu\Delta\boldsymbol u
=
\boldsymbol 0,
\]

together with

\[
\nabla\cdot\boldsymbol u=0,
\]

where

\[
\nu=0.01.
\]

The computational domain is

\[
\Omega
=
([0,4]\times[0,2])
\setminus
([0,2]\times[1,2]).
\]

At the inlet \(x=0\), \(0\le y\le1\),

\[
u(0,y)=4y(1-y),
\qquad
v(0,y)=0.
\]

At the outlet \(x=4\),

\[
p(4,y)=0.
\]

The no-slip condition is imposed on the remaining solid walls:

\[
u=v=0.
\]

\subsection{Lid-driven cavity flow (\texttt{navier\_stokes\_2d\_classic})}\label{d.12-lid-driven-cavity-flow-navier_stokes_2d_classic}

This problem corresponds to the benchmark case \texttt{navier\_stokes\_2d\_C}.

The steady incompressible Navier--Stokes equations are

\[
(\boldsymbol u\cdot\nabla)\boldsymbol u
+
\nabla p
-
\nu\Delta\boldsymbol u
=
\boldsymbol 0,
\]

and

\[
\nabla\cdot\boldsymbol u=0,
\]

on

\[
(x,y)\in[0,1]^2.
\]

The viscosity is

\[
\nu=0.01,
\]

corresponding to

\[
\mathrm{Re}=100.
\]

On the moving top wall,

\[
u(x,1)=4x(1-x),
\qquad
v(x,1)=0.
\]

On the other three walls,

\[
u=v=0.
\]

Pressure is fixed by the gauge condition

\[
p(0,0)=0.
\]

\subsection{Long-time incompressible Navier--Stokes equation (\texttt{navier\_stokes\_2d\_long\_time})}\label{d.13-long-time-incompressible-navierstokes-equation-navier_stokes_2d_long_time}

The governing equations are

\[
\boldsymbol u_t
+
(\boldsymbol u\cdot\nabla)\boldsymbol u
+
\nabla p
-
\nu\Delta\boldsymbol u
=
\boldsymbol f(x,y,t),
\]

and

\[
\nabla\cdot\boldsymbol u=0,
\]

on

\[
(x,y,t)\in[0,2]\times[0,1]\times[0,5],
\]

with

\[
\nu=0.01.
\]

The forcing term is

\[
\boldsymbol f(x,y,t)
=
\begin{pmatrix}
0\\
-\sin(\pi x)\sin(\pi y)\sin(\pi t)
\end{pmatrix}.
\]

At the inlet \(x=0\),

\[
u(0,y,t)
=
\sin(\pi y)
\left[
\sin(\pi t)
+
\sin(3\pi t)
+
\sin(5\pi t)
\right],
\]

and

\[
v(0,y,t)=0.
\]

At the outlet \(x=2\),

\[
p(2,y,t)=0.
\]

On the remaining spatial boundaries,

\[
u=v=0.
\]

The initial condition is

\[
u(x,y,0)=v(x,y,0)=p(x,y,0)=0.
\]

\subsection{Classical Poisson equation on a perforated domain (\texttt{poisson\_2d\_classic})}\label{d.14-classical-poisson-equation-on-a-perforated-domain-poisson_2d_classic}

The governing equation is

\[
\Delta u=0
\qquad
\text{in }\Omega,
\]

where

\[
\Omega
=
\left[-\frac12,\frac12\right]^2
\setminus
\bigcup_{\boldsymbol c\in\mathcal C}
B_{0.1}(\boldsymbol c),
\]

and

\[
\mathcal C
=
\{
(0.3,0.3),
(-0.3,0.3),
(0.3,-0.3),
(-0.3,-0.3)
\}.
\]

The Dirichlet boundary conditions are

\[
u=1
\qquad
\text{on the outer rectangular boundary},
\]

and

\[
u=0
\qquad
\text{on the four circular boundaries}.
\]

\subsection{Screened Poisson--Boltzmann problem (\texttt{poisson\_boltzmann\_2d})}\label{d.15-screened-poissonboltzmann-problem-poisson_boltzmann_2d}

The benchmark considers the screened Poisson equation

\[
-\Delta u+k^2u=f(x,y),
\]

with

\[
k=8.
\]

The source term is

\[
f(x,y)
=
A
\left(
\mu_1^2+x^2+\mu_2^2+y^2
\right)
\sin(\mu_1\pi x)
\sin(\mu_2\pi y),
\]

where

\[
A=10,
\qquad
\mu_1=1,
\qquad
\mu_2=4.
\]

The spatial domain is

\[
\Omega
=
[-1,1]^2
\setminus
\bigcup_{j=1}^{4}
B_{r_j}(\boldsymbol c_j),
\]

where

\[
\begin{aligned}
(\boldsymbol c_1,r_1)&=((0.5,0.5),0.2),\\
(\boldsymbol c_2,r_2)&=((0.4,-0.4),0.4),\\
(\boldsymbol c_3,r_3)&=((-0.2,-0.7),0.1),\\
(\boldsymbol c_4,r_4)&=((-0.6,0.5),0.3).
\end{aligned}
\]

The Dirichlet conditions are

\[
u=0.2
\qquad
\text{on the outer boundary},
\]

and

\[
u=1
\qquad
\text{on the circular boundaries}.
\]

\subsection{Multiscale multi-area Poisson equation (\texttt{poisson\_2d\_many\_area})}\label{d.16-multiscale-multi-area-poisson-equation-poisson_2d_many_area}

The governing equation is

\[
a(x,y)\Delta u+f(x,y)=0,
\qquad
(x,y)\in[-10,10]^2.
\]

The domain is partitioned into \(5\times5\) rectangular subdomains. Within subdomain \(\Omega_{ij}\),

\[
a(x,y)=a_{ij},
\]

where \(a_{ij}\) is read from \texttt{poisson\_a\_coef.dat}.

The local forcing is defined by the tabulated coefficients in \texttt{poisson\_f\_coef.dat}:

\[
f(x,y)
=
c_{ij,00}
+
\sum_{r=0}^{1}
\sum_{s=0}^{1}
c_{ij,rs}
\sin\left(
\frac{\pi r\,\xi}{h_x}
\right)
\sin\left(
\frac{\pi s\,\eta}{h_y}
\right),
\qquad
(x,y)\in\Omega_{ij},
\]

where \((\xi,\eta)\) denote the local coordinates within \(\Omega_{ij}\).

The Robin boundary condition is

\[
\partial_n u=-u
\qquad
\text{on }\partial[-10,10]^2.
\]

\subsection{Three-dimensional Poisson equation on a complex geometry (\texttt{poisson\_3d\_complex\_geometry})}\label{d.17-three-dimensional-poisson-equation-on-a-complex-geometry-poisson_3d_complex_geometry}

The governing equation is

\[
-\mu(z)\Delta u
+
\kappa(z)u
=
f(x,y,z),
\]

in a unit cube with four spherical holes:

\[
\Omega
=
[0,1]^3
\setminus
\bigcup_{j=1}^{4}
B_{r_j}(\boldsymbol c_j),
\]

where

\[
\begin{aligned}
(\boldsymbol c_1,r_1)&=((0.4,0.3,0.6),0.2),\\
(\boldsymbol c_2,r_2)&=((0.6,0.7,0.6),0.2),\\
(\boldsymbol c_3,r_3)&=((0.2,0.8,0.7),0.1),\\
(\boldsymbol c_4,r_4)&=((0.6,0.2,0.3),0.1).
\end{aligned}
\]

The coefficients are

\[
\mu(z)=1,
\]

and

\[
\kappa(z)
=
\begin{cases}
8^2, & z<0.5,\\
10^2, & z\ge0.5.
\end{cases}
\]

Let

\[
r^2=x^2+y^2+z^2.
\]

The forcing term is

\[
\begin{aligned}
f(x,y,z)
={}&
20
\exp\left[
\sin(\pi x)
+
\sin(10\pi y)
+
\sin(5\pi z)
\right]
\frac{r^2-1}{r^2+1}\\
&+
100
\sin(\pi x)
\sin(10\pi y)
\sin(5\pi z).
\end{aligned}
\]

A homogeneous Neumann condition is imposed on the complete boundary:

\[
\partial_n u=0
\qquad
\text{on }\partial\Omega.
\]

\subsection{Five-dimensional Poisson equation (\texttt{poisson\_5d})}\label{d.18-five-dimensional-poisson-equation-poisson_5d}

The governing equation is

\[
-\Delta u
=
\frac{\pi^2}{4}
\sum_{i=1}^{5}
\sin\left(
\frac{\pi x_i}{2}
\right),
\qquad
\boldsymbol x\in[0,1]^5.
\]

The analytical reference solution is

\[
u^\star(\boldsymbol x)
=
\sum_{i=1}^{5}
\sin\left(
\frac{\pi x_i}{2}
\right).
\]

The Dirichlet boundary condition is

\[
u=u^\star
\qquad
\text{on }\partial[0,1]^5.
\]

\subsection{Inverse Poisson equation (\texttt{poisson\_inverse\_2d})}\label{d.19-inverse-poisson-equation-poisson_inverse_2d}

The unknowns are \(u(x,y)\) and the spatially varying diffusion coefficient \(a(x,y)\), satisfying

\[
-\nabla\cdot(a\nabla u)
=
f(x,y),
\qquad
(x,y)\in[0,1]^2.
\]

The manufactured reference fields are

\[
u^\star(x,y)
=
\sin(\pi x)\sin(\pi y),
\]

and

\[
a^\star(x,y)
=
\frac{
1
}{
1+x^2+y^2+(x-1)^2+(y-1)^2
}.
\]

Writing

\[
a=a^\star(x,y),
\]

the corresponding source term is

\[
\begin{aligned}
f(x,y)
={}&
2\pi^2a
\sin(\pi x)\sin(\pi y)\\
&+
2\pi a^2
\Big[
(2x-1)\cos(\pi x)\sin(\pi y)\\
&\hspace{18mm}
+
(2y-1)\sin(\pi x)\cos(\pi y)
\Big].
\end{aligned}
\]

The diffusion coefficient is prescribed on the spatial boundary:

\[
a=a^\star
\qquad
\text{on }\partial[0,1]^2.
\]

The inverse problem uses a \(50\times50\) grid of noisy state observations:

\[
u_{\mathrm{obs}}
=
u^\star+\epsilon,
\qquad
\epsilon\sim\mathcal N(0,0.1^2).
\]

\subsection{One-dimensional wave equation (\texttt{wave\_1d})}\label{d.20-one-dimensional-wave-equation-wave_1d}

The governing equation is

\[
u_{tt}-4u_{xx}=0,
\qquad
(x,t)\in[0,1]\times[0,1].
\]

The initial conditions are

\[
u(x,0)
=
\sin(\pi x)
+
\frac12\sin(4\pi x),
\]

and

\[
u_t(x,0)=0.
\]

The spatial boundary condition is

\[
u(0,t)=u(1,t)=0.
\]

The analytical reference solution is

\[
u^\star(x,t)
=
\sin(\pi x)\cos(2\pi t)
+
\frac12
\sin(4\pi x)\cos(8\pi t).
\]

\subsection{Heterogeneous two-dimensional wave equation (\texttt{wave\_2d\_heterogeneous})}\label{d.21-heterogeneous-two-dimensional-wave-equation-wave_2d_heterogeneous}

The governing equation is

\[
\Delta u
-
\frac{1}{c(x,y)}u_{tt}
=
0,
\]

or equivalently,

\[
u_{tt}
=
c(x,y)\Delta u,
\]

on

\[
(x,y,t)\in[-1,1]^2\times[0,5].
\]

The heterogeneous coefficient \(c(x,y)\) is defined by interpolation of the Gaussian-random-field realization stored in \texttt{darcy\_2d\_coef\_256.dat}.

The initial conditions are

\[
u(x,y,0)
=
\exp\left[
-\frac{(x+0.5)^2+y^2}{2(0.3)^2}
\right],
\]

and

\[
u_t(x,y,0)=0.
\]

A homogeneous Neumann condition is imposed on the spatial boundary:

\[
\partial_n u=0
\qquad
\text{on }\partial[-1,1]^2.
\]

\subsection{One-dimensional Allen--Cahn equation (\texttt{allen\_cahn\_1d})}\label{d.22-one-dimensional-allencahn-equation-allen_cahn_1d}

The governing equation is

\[
u_t
-
\varepsilon^2u_{xx}
+
u^3-u
=
0,
\]

on

\[
(x,t)\in[-1,1]\times[0,0.25],
\]

with

\[
\varepsilon=0.1.
\]

The initial condition is

\[
u(x,0)
=
\tanh\left(
\frac{x}{\sqrt{2}\varepsilon}
\right).
\]

The Dirichlet boundary condition is obtained from the stationary reference solution:

\[
u(\pm1,t)
=
\tanh\left(
\frac{\pm1}{\sqrt{2}\varepsilon}
\right).
\]

The analytical reference solution is

\[
u^\star(x,t)
=
\tanh\left(
\frac{x}{\sqrt{2}\varepsilon}
\right).
\]

\subsection{Variable-coefficient Darcy equation (\texttt{darcy\_flow\_2d})}\label{d.23-variable-coefficient-darcy-equation-darcy_flow_2d}

The governing equation is

\[
-\nabla\cdot
\left(
a(x,y)\nabla u
\right)
=
f(x,y),
\qquad
(x,y)\in[0,1]^2.
\]

The heterogeneous coefficient is

\[
a(x,y)
=
2
+
\frac12
\sin(2\pi x)
\sin(3\pi y).
\]

The manufactured reference solution is

\[
u^\star(x,y)
=
\sin(\pi x)\sin(\pi y).
\]

The source term is defined consistently as

\[
f(x,y)
=
-\nabla\cdot
\left(
a(x,y)\nabla u^\star(x,y)
\right).
\]

Equivalently,

\[
f
=
-
\left(
a_xu_x^\star
+
a_yu_y^\star
+
a\Delta u^\star
\right).
\]

The homogeneous Dirichlet boundary condition is

\[
u=0
\qquad
\text{on }\partial[0,1]^2.
\]

\subsection{Two-dimensional shallow-water system (\texttt{shallow\_water\_2d})}\label{d.24-two-dimensional-shallow-water-system-shallow_water_2d}

Let \(h\) denote the water depth, and let \(q_x\) and \(q_y\) denote the two momentum components.

The conservative shallow-water equations are

\[
h_t+(q_x)_x+(q_y)_y=0,
\]

\[
(q_x)_t
+
\left(
\frac{q_x^2}{h}
+
\frac{gh^2}{2}
\right)_x
+
\left(
\frac{q_xq_y}{h}
\right)_y
=
0,
\]

and

\[
(q_y)_t
+
\left(
\frac{q_xq_y}{h}
\right)_x
+
\left(
\frac{q_y^2}{h}
+
\frac{gh^2}{2}
\right)_y
=
0,
\]

on

\[
(x,y,t)\in[0,1]^2\times[0,0.1],
\]

with gravitational acceleration

\[
g=9.81.
\]

The initial state is spatially uniform:

\[
(h,q_x,q_y)(x,y,0)
=
(1,0.1,-0.05).
\]

All three variables are periodic in both spatial directions:

\[
\boldsymbol U(0,y,t)
=
\boldsymbol U(1,y,t),
\]

and

\[
\boldsymbol U(x,0,t)
=
\boldsymbol U(x,1,t),
\]

where

\[
\boldsymbol U
=
(h,q_x,q_y)^\mathsf{T}.
\]

Water-depth positivity is enforced through

\[
h\ge h_{\min},
\qquad
h_{\min}=10^{-3}.
\]

The reference solution is the constant state

\[
\boldsymbol U^\star
=
(1,0.1,-0.05)^\mathsf{T}.
\]

\subsection{Kovasznay flow (\texttt{kovasznay\_flow\_2d})}\label{d.25-kovasznay-flow-kovasznay_flow_2d}

The steady incompressible Navier--Stokes equations are

\[
(\boldsymbol u\cdot\nabla)\boldsymbol u
+
\nabla p
-
\nu\Delta\boldsymbol u
=
\boldsymbol 0,
\]

together with

\[
\nabla\cdot\boldsymbol u=0.
\]

The computational domain is

\[
(x,y)\in[-0.5,1]\times[-0.5,1.5].
\]

The Reynolds number and viscosity are

\[
\mathrm{Re}=40,
\qquad
\nu=\frac{1}{40}.
\]

Define

\[
\lambda
=
\frac{\mathrm{Re}}{2}
-
\sqrt{
\frac{\mathrm{Re}^2}{4}
+
4\pi^2
}.
\]

The analytical reference solution is

\[
u^\star(x,y)
=
1-e^{\lambda x}\cos(2\pi y),
\]

\[
v^\star(x,y)
=
\frac{\lambda}{2\pi}
e^{\lambda x}
\sin(2\pi y),
\]

and

\[
p^\star(x,y)
=
\frac12
\left(
1-e^{2\lambda x}
\right).
\]

The analytical values are imposed on the rectangular boundary:

\[
(u,v,p)
=
(u^\star,v^\star,p^\star)
\qquad
\text{on }\partial\Omega.
\]

Pressure is additionally anchored by

\[
p(0,0)=0.
\]

\subsection{Note on the excluded PINNacle Wave2d-MS problem}\label{d.26-note-on-the-excluded-pinnacle-wave2d-ms-problem}

Among the PINNacle problems, we exclude the 2D multi-scale long-time wave equation (Wave2d-MS) because its published mathematical specification is internally inconsistent. PINNacle defines the governing equation as

\[
u_{tt}-(u_{xx}+a^2u_{yy})=0,
\qquad
a=\sqrt{2},
\]

while the reported reference solution is

\[
\sinh(\pi x)\sinh(\pi y)
\cos(\sqrt{3}\pi t).
\]

Direct substitution gives

\[
-6\pi^2u^\star
\neq 0.
\]

showing that the reported reference solution does not satisfy the stated governing equation.

We further observed that the reference solution in the released PINNacle implementation differs from the one reported in the paper and, under the released default parameters, is also inconsistent with the stated PDE. To avoid evaluating PINN methods against an ambiguous PDE--reference pair, we therefore exclude Wave2d-MS from our benchmark rather than modifying its original definition. All other PINNacle problems used in this work follow the corresponding benchmark formulations described above.

\section{Broader Impact and Limitations}\label{appendix-e-broader-impact-limitations}

PINNForge aims to reduce manual trial and error while making automated PINN design more traceable through structured configurations and execution evidence. Lower reference error, however, does not guarantee physical correctness or generalization. The same fixed reference points are reused for search control, final selection, and reporting, so the reported MSE measures this benchmark protocol rather than held-out generalization; adaptive reuse may favor configurations tuned to those points. Designed solvers may also fail in unsampled regions, out-of-distribution regimes, or unseen parameter settings. Independent evaluation points, physical-consistency checks, and domain-expert assessment would provide stronger validation.

The experiments use a single LLM backbone, Qwen3.8-27B, and no provider-side sampling seed, so they do not establish cross-backbone robustness or exact replay of individual search trajectories. PINNForge also incurs LLM inference overhead through multi-round generation and analysis. The optimizer-step comparison in Section 4.3 therefore measures PINN training cost rather than total computational or monetary cost. Future evaluation should consider benchmark accuracy together with LLM inference, wall-clock time, and overall resource use.